\documentclass{article}

\usepackage[nonatbib,final]{neurips_2022}

\usepackage[utf8]{inputenc} 
\usepackage[T1]{fontenc}    
\usepackage{hyperref}       
\usepackage{url}            
\usepackage{booktabs}       
\usepackage{amsfonts}       
\usepackage{nicefrac}       
\usepackage{microtype}      
\usepackage{xcolor}         
\usepackage{subfigure}
\usepackage{multirow}
\usepackage{xspace}
\usepackage{listings}
\usepackage{xcolor}
\usepackage{amsmath,amssymb}
\usepackage[linesnumbered,vlined,ruled,noend]{algorithm2e}
\usepackage{wrapfig}
\usepackage{enumitem}
\usepackage{CJKutf8}
\usepackage{floatrow}
\usepackage{bm}
\newfloatcommand{capbtabbox}{table}[][\FBwidth]
\usepackage{graphicx}
\usepackage{amsmath,amsfonts,graphicx}
\usepackage{amsthm}

\usepackage{makecell}
\usepackage{CJKutf8}
\usepackage{caption}
\newcommand{\sys}{DivBO\xspace}

\title{\sys: Diversity-aware CASH for Ensemble Learning}

\author{
Yu Shen$^1$, Yupeng Lu$^1$, Yang Li$^4$, Yaofeng Tu$^3$, Wentao Zhang$^5$$^6$, Bin Cui$^1$$^2$ \\
$^1$Key Lab of High Confidence Software Technologies, Peking University, China \\
$^2$Institute of Computational Social Science, Peking University (Qingdao), China\\
$^3$ ZTE Corporation, China $^4$ Data Platform, TEG, Tencent Inc., China\\
$^5$ Mila - Québec AI Institute $^6$HEC, Montréal, Canada\\
$^1$\{shenyu, xinkelyp, bin.cui\}@pku.edu.cn, 
$^2$ tu.yaofeng@zte.com.cn\\
$^3$thomasyngli@tencent.com
$^4$wentao.zhang@mila.quebec\\
}

\begin{document}

\maketitle

\begin{abstract}
The Combined Algorithm Selection and Hyperparameters optimization (CASH) problem is one of the fundamental problems in Automated Machine Learning (AutoML). 
Motivated by the success of ensemble learning, recent AutoML systems build post-hoc ensembles to output the final predictions instead of using the best single learner.
However, while most CASH methods focus on searching for a single learner with the best performance, they neglect the diversity among base learners (i.e., they may suggest similar configurations to previously evaluated ones), which is also a crucial consideration when building an ensemble.
To tackle this issue and further enhance the ensemble performance, we propose \sys, a diversity-aware framework to inject explicit search of diversity into the CASH problems.
In the framework, we propose to use a diversity surrogate to predict the pair-wise diversity of two unseen configurations.
Furthermore, we introduce a temporary pool and a weighted acquisition function to guide the search of both performance and diversity based on Bayesian optimization.
Empirical results on 15 public datasets show that \sys achieves the best average ranks (1.82 and 1.73) on both validation and test errors among 10 compared methods, including post-hoc designs in recent AutoML systems and state-of-the-art baselines for ensemble learning on CASH problems.

\end{abstract}

\section{Introduction}
\label{sec:intro}
In recent years, machine learning has made great strides in various application areas, e.g., computer vision~\cite{he2016deep,goodfellow2014generative}, recommendation~\cite{su2009survey,sun2019bert4rec}, etc.
However, it's often knowledge-intensive to develop customized solutions with promising performance, as the process includes selecting proper ML algorithms and tuning the hyperparameters. 
To reduce the barrier and facilitate the deployment of machine learning applications, the AutoML community raises the Combined Algorithm Selection and Hyperparameters optimization (CASH) problem~\cite{thornton2013auto} and proposes several methods~\cite{quanming2018taking,hutter2019automated,he2021automl} to automate the optimization.

While most AutoML methods~\cite{snoek2012practical,hutter2011sequential,bergstra2011algorithms} for CASH focus on searching for the optimal performance of a single learner, it's widely acknowledged that ensembles of promising learners often outperform single ones~\cite{zhang2020efficient,bian2021does}. 
For example, He et al.~\cite{he2016deep} won first place in ILSRVC 2015 with an average of several learners. 
And ensemble strategies can be frequently found in the top solutions of Kaggle competitions~\cite{hoch2015ensemble,bojer2021kaggle}.
Motivated by those achievements, recent AutoML systems (e.g., Auto-sklearn~\cite{feurer2015efficient}, Auto-Pytorch~\cite{zimmer2021auto}, VolcanoML~\cite{li2021volcanoml}) build post-hoc ensembles based on all base learners from the entire optimization process and show better empirical results than using the best single learner.

\setlength{\intextsep}{0pt}%
\begin{wrapfigure}[12]{R}{0.465\textwidth}
  \begin{center}
    \includegraphics[width=1\textwidth]{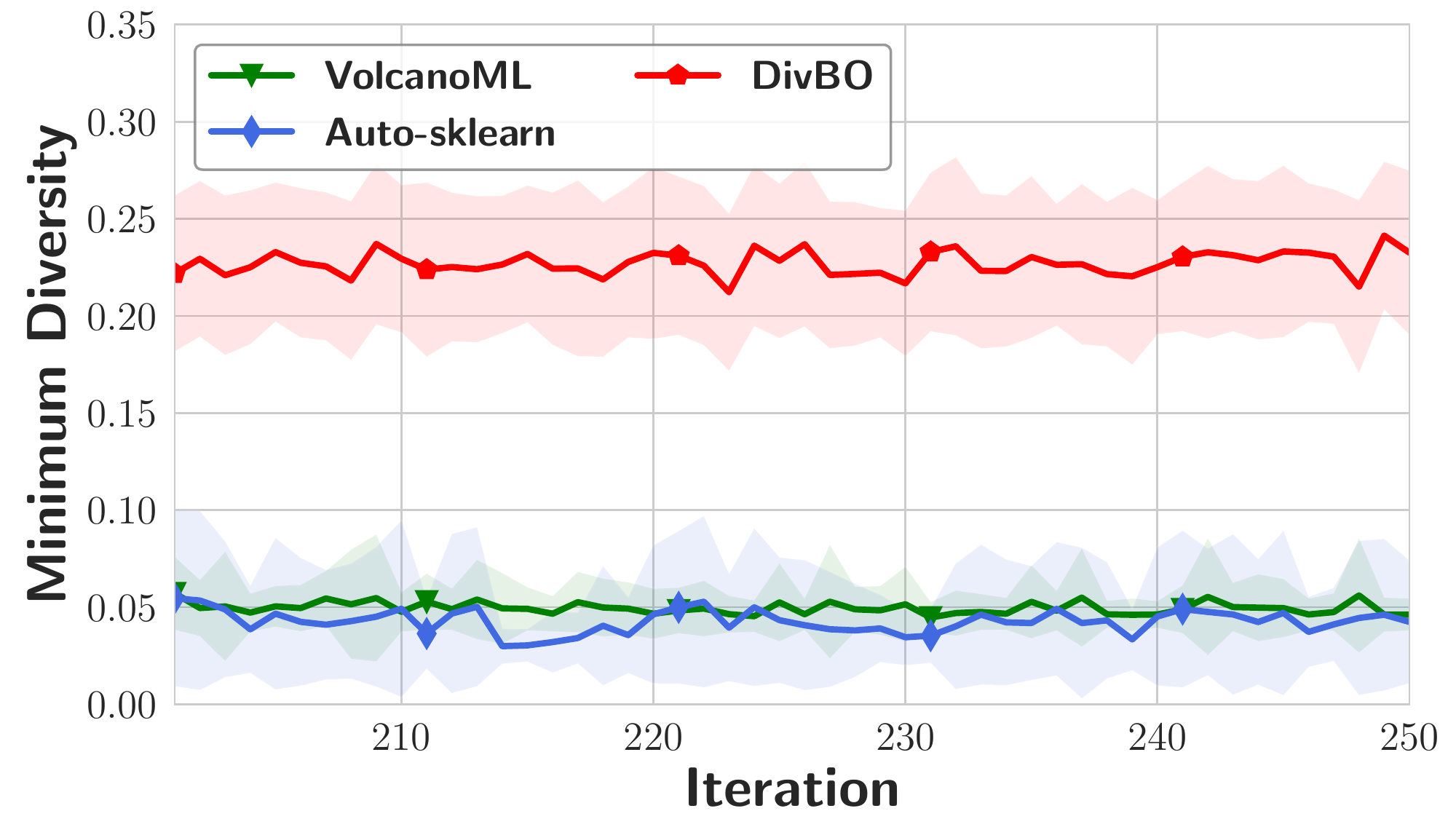}
  \end{center}
  \vspace{-2mm}
  \caption{Minimum diversity of suggested learners after 200 iterations on quake. (20 runs)}
  \label{fig:intro}
\end{wrapfigure}

Despite the effectiveness of those post-hoc ensemble designs, the target of CASH methods is inconsistent with that of ensemble learning. 
In other words, a good ensemble should contain a pool of base learners that are both well-performing and diverse with each other~\cite{zaidi2021neural,zhang2020efficient}, while most CASH methods only aim at searching for the best-performing configuration of learners. Figure~\ref{fig:intro} shows the diversity of the suggested learners from Auto-sklearn and VolcanoML after 200 iterations.
We define the pair-wise diversity of two learners as the average regularized Euclidean distance of their probabilistic predictions which ranges from 0 to 1 (See Equation~\ref{eq:div}). 
And we show the minimum diversity of a suggested learner with other learners from the ensemble built at the 200$^{th}$ iteration.
While most diversity values of the two AutoML systems are around 0.05, they suggest learners whose predictions are often quite similar to one of the learners obtained from previous iterations. 
It implies a lack of diversity in previous post-hoc ensemble designs, and it also indicates that the performance of post-hoc ensembles might be further improved if diversity is also taken into consideration when suggesting new configurations to evaluate.
However, the search of diversity is non-trivial. 
A simple diversity-inducing algorithm would degenerate the pool of learners, e.g., diversity can be easily increased by predicting all samples for the classes that are wrong and different from previous learners.
Therefore, how to guide the search for both performance and diversity simultaneously in recent CASH algorithms is still an open question.

In this paper, we propose \sys, a new algorithm framework that combines Bayesian optimization (BO) with an explicit search of diversity for classification. 
The contributions are summarized as:



1. To the best of our knowledge, this work is the first to enhance the ensemble performance of CASH methods by considering the diversity of suggested configurations. 

2. To inject the search of diversity into CASH problems, we design 1) the diversity surrogate to predict the pair-wise diversity given two unseen configurations and 2) a BO-based framework that automatically balances the predictive performance and diversity of suggestions.

3. Empirical results show that the diversity surrogate achieves a better correlation with the ground-truth results compared with the performance surrogate used in BO. 
Compared with recent AutoML systems, \sys suggests configurations that are significantly more diverse with those in previous iterations (the red line in Figure~\ref{fig:intro}).
In addition, results on 15 public datasets show that \sys achieves the best average rank (1.82 and 1.73) on both validation and test errors among 10 methods, including post-hoc designs in recent AutoML systems and state-of-the-art methods for ensemble learning on CASH problems.

\section{Preliminary and Related Work}
In this section, we review the related work and introduce the preliminary for our proposed method.

\textbf{Combined algorithm selection and hyperparameters (CASH).}
We first introduce the basic notations for the CASH problem.
Let $\mathcal{A}$ be the set of $K$ candidate algorithms $ \mathcal{A} = \{A^1, ..., A^K\}$. 
Each algorithm $A^i$ has its corresponding hyperparameter space $\Lambda_{A^i}$.
Given dataset $\mathcal{D}=\{\mathcal{D}_{train},\mathcal{D}_{val}\}$ of a learning problem, the CASH problem aims to minimize the validation metric $\mathcal{L}$ by searching for the best joint configuration $x^*= (A^*,\lambda^*)$ trained on the training set, where $A^*$ is the best algorithm and $\lambda^*$ is its corresponding hyperparameters.
For brevity, we use $T_{(a,\lambda)}$ to denote the learner constructed by the joint configuration $x=(a, \lambda)$ and trained on $\mathcal{D}_{train}$.
The optimization objective can be formulated as:
\begin{equation}
\label{eq:cash}
    \min_{a \in \mathcal{A}, \lambda \in \Lambda_a} \mathcal{L}(T_{(a,\lambda)}, \mathcal{D}_{val}).
\end{equation}
The CASH problem is first introduced by Auto-WEKA~\cite{thornton2013auto} and solved by applying Bayesian optimization (BO)~\cite{snoek2012practical,hutter2011sequential,bergstra2011algorithms} on the entire search space.
Each BO iteration follows three steps: 
1) BO fits a probabilistic surrogate model based on observations $D=\{(x_1, y_1),...,(x_{n-1}, y_{n-1})\}$, in which $x_i$ is the $i^{th}$ evaluated joint configuration, and $y_i$ is its corresponding observed performance;
2) BO selects the most promising configuration $x_n$ by maximizing the acquisition function to balance exploration and exploitation; 
3) BO evaluates the configuration $x_n$ (i.e., train the learner and obtain its validation performance), and augment the observations $D$. 
To avoid confusion, we refer to the surrogate of BO as the performance surrogate $M_{perf}$ in the following section.

In addition to BO, TPOT~\cite{olson2016tpot} proposes to use genetic programming to solve the CASH problem. 
The ADMM-based method~\cite{liu2020admm} decomposes the problem into sub-problems and solves them using ADMM~\cite{boyd2011distributed}. 
Rising bandit~\cite{li2020efficient} continuously eliminates badly performing algorithms during optimization.
FLAML~\cite{wang2021flaml} suggests configurations based on estimated costs of improvement.

\textbf{Ensemble-oriented CASH.}
Different from the original CASH, ensemble-oriented CASH aims to find an ensemble of $m$ learners that minimizes the validation metric $\mathcal{L}$, which is:
\begin{equation}
\label{eq:cash_es}
    \min_{a_1,...,a_m \in \mathcal{A}, \lambda_1 \in \Lambda_{a_1},...,\lambda_m \in \Lambda_{a_m}} \mathcal{L}(Ensemble(T_{(a_1,\lambda_1)},...,T_{(a_m,\lambda_m)}),\mathcal{D}_{val}).
\end{equation}
To solve the problem, ensemble optimization~\cite{levesque2016bayesian} directly optimizes the target based on BO by using an ensemble pool with a fixed size. 
This strategy suffers from instability due to the risk of adding a bad configuration during optimization.
Neural ensemble search (NES)~\cite{zaidi2021neural} is proposed based on the regularized evolutionary algorithm~\cite{real2019regularized}.
The local search design in NES works well on low-dimensional problems like neural architecture search, but may encounter a bottleneck when applied to solve high-dimensional CASH problems.

Rather than directly optimizing on Equation~\ref{eq:cash_es}, recent AutoML systems (e.g., Auto-sklearn~\cite{feurer2015efficient}, VolcanoML~\cite{li2021volcanoml,li2022volcanoml}, Auto-Pytorch~\cite{zimmer2021auto}) adopt a specific ensemble strategy on the observation history after solving Equation~\ref{eq:cash}, which we refer to as the \textbf{post-hoc ensemble} designs. 
While the optimization target is inconsistent, as shown in Section~\ref{sec:intro}, there's still space to improve the performance of post-hoc ensembles.
Inspired by the theoretical study that diversity can help improve the performance of the ensemble~\cite{lecun2015deep,bian2021does,hansen1990neural,zhou2002ensembling}, we follow the post-hoc ensemble designs and focus on generating a promising pool of learners for the final ensemble strategy.

\textbf{Ensemble selection.}
Ensemble strategies are methods that combine the predictions given a pool of learners, which are orthogonal to the direction of our method (i.e., suggesting a diverse pool of learners).
Among different ensemble strategies (e.g., Bagging~\cite{dietterich2000ensemble}, Boosting~\cite{dietterich2000ensemble,moghimi2016boosted}, Stacking~\cite{breiman1996stacked}), we adopt the ensemble selection~\cite{caruana2004ensemble}, which works empirically well with AutoML as shown in previous study~\cite{feurer2015efficient,li2021volcanoml,zaidi2021neural}.
In short, ensemble selection starts from an empty ensemble and iteratively adds models from the pool with replacement to maximize the ensemble validation performance (with uniform weights).
The pseudo-code is provided in Appendix A.1.

\section{Diversity-aware Bayesian optimization (\sys)}
In this section, we present \sys \xspace --- our proposed diversity-aware CASH method for ensemble learning based on Bayesian optimization (BO). To inject the search of diversity into BO, we will answer the following two questions: 1) how to measure diversity and model the diversity of two unseen configurations, and 2) how to suggest configurations that are both well-performing and diverse with potential ones in the final ensemble.

\subsection{Diversity Surrogate}
\label{sec:surrogate}

The existing diversity measures can be generally divided into pair-wise and nonpair-wise measures. The nonpair-wise diversity~\cite{zhou2012ensemble} directly measures a set of learners in the ensemble. 
While the ensemble may change during each iteration, it’s difficult to model the diversity of an ensemble with a candidate configuration and multiple learners. 
Therefore, we use the pair-wise measures to simply learn the diversity of two given configurations. 
To this end, we follow the definition in previous research~\cite{zhang2020efficient} that explicitly improves the diversity of neural networks and shows satisfactory empirical results.
Let $x_i$ denote as the joint configuration of algorithm $a_i$ and hyperparameters $\lambda_i$. 
The diversity between two configurations $(x_i,x_j)$ is the average Euclidean distance of class probabilities predicted on the validation set~\cite{zhang2020efficient}, which is:
\begin{equation}
    Div(x_i,x_j)=\frac{\sqrt{2}}{2}\frac{1}{|\mathcal{D}_{val}|}\sum_{s \in \mathcal{D}_{val}}|| T_{x_i}(s)-T_{x_j}(s) ||_2,
    \label{eq:div}
\end{equation}
where $\mathcal{D}_{val}$ is the validation set, $T_{x_i}$ is the learner corresponding to configuration $x_i$ and fitted on the training set, and $T_{x_i}(s)$ is the predictive class probability on sample $s$.
Obviously, the relationship between the diversity and configuration pair $(x_i,x_j)$ is also a black-box function. 
Therefore, inspired by Bayesian optimization, we apply another surrogate to model this relationship. 
This surrogate, namely the diversity surrogate, takes a configuration pair as input and outputs the predictive mean and variance of the pair-wise diversity.

\textbf{Fitting.} 
During each iteration, \sys generates the training set for the diversity surrogate $M_{div}$ by computing the diversity value of each pair of the observed configurations. 
Note that, $(x_i,x_j)$ and $(x_j,x_i)$ lead to the same diversity value, but the training set should include both of them to ensure symmetry.
This leads to about $|D|^2$ training samples, where $|D|$ is the number of observations.
To avoid fetching validation predictions by retraining, we store those predictions for all observations during optimization, which is a common trick applied in previous methods~\cite{feurer2015efficient,levesque2016bayesian,zaidi2021neural}.

\textbf{Implementation.}
Concretely, the diversity surrogate $M_{div}$ in \sys is an ensemble of several LightGBM models~\cite{ke2017lightgbm}. This implementation has the following two advantages:
1) The diversity surrogate gives predictions with uncertainty to balance exploration and exploitation. Concretely, the predictive mean and variance of the surrogate is obtained by computing the mean and variance of outputs generated by different LightGBM models.
2) The time complexity of fitting the surrogate is relatively low. 
The cost of training a LightGBM model is $O(|D|^2\log|D|)$, which is lower than $O(|D|^3)$ when fitting a Gaussian Process as the performance surrogate.
In Section~\ref{sec:exp}, we will further compare different implementations of diversity surrogates on their ability to fit the relationship. 

\subsection{Diversity-aware Framework}
\label{sec:framework}
Based on the diversity surrogate, we propose a diversity-aware framework to suggest configurations that are both well-performing and diverse. 
Before stepping into the design, consider a straightforward strategy that we choose the most diverse configuration with the observation history during each iteration. 
This simple strategy has two obvious drawbacks: 
1) First, there is no need to suggest configurations that are diverse from badly-performing ones in the observation history---we only expect the promising base learners in the final ensemble to be diverse with each other; 
2) The diversity can be easily increased by suggesting a learner that predicts all samples for the classes that are wrong and different from previous learners. 
In this extreme case, the learner can not help improve the performance of the final ensemble.
Therefore, how to control the diversity of the suggested configurations is non-trivial. 
In the following, we explain how \sys tackles the two drawbacks.

\textbf{Suggesting diverse configurations.} 
To tackle the first drawback, since which learners will appear in the final ensemble is unknown during optimization,
\sys proposes to use a \textbf{temporary pool} $\mathcal{P}$ to collect base learners that will probably be selected into the final ensemble. 
Concretely, the pool is built by applying ensemble selection to the observation history. 
When the optimization ends, this temporary pool is exactly the output of a post-hoc ensemble strategy (e.g., auto-sklearn\cite{feurer2015efficient}) that ends at the previous iteration, 
and thus the configurations in the temporary pool can be regarded as the potential ones in the final ensemble. 
Then, we define the diversity acquisition function $\alpha_{div}$ of each unseen configuration $x$ as its predicted diversity value with the most similar configuration in the temporary pool:
\begin{equation}
    \alpha_{div}(x)=\frac{1}{N}\sum_{n=1}^N\min_{\theta \in \mathcal{P}} M^n_{div}(\theta,x),
\label{eq:div_acq}
\end{equation}
where $\mathcal{P}$ is the temporary configuration pool and $M_{div}$ is the diversity surrogate. 
$M^n_{div}(\theta,x)$ is the $n^{th}$ sampled value from the output distribution of $M_{div}$ given the pair $(\theta,x)$, and the final acquisition function is the average of $N$ minimums via sampling.
Note that, using the minimum diversity with learners in the temporary pool is more appropriate than the mean diversity. 
In extreme cases, when the algorithm may suggest a useless configuration that is exactly the same as a previously evaluated one, the minimum diversity is penalized to zero while the mean diversity is still larger than zero.
By maximizing Equation~\ref{eq:div_acq}, \sys is able to suggest configurations that are diverse from other potential ones in the final ensemble during optimization.

\textbf{Combining predictive performance and diversity.}
As mentioned in the second drawback, optimizing diversity alone degenerates the predictive performance of the suggested configurations.
\sys tackles this issue by combining both the performance and diversity acquisition functions with a saturating weight. 
Since the performance and diversity acquisition values are of different scales, we propose to use the sum of ranking values instead of directly adding the output values. 
During each BO iteration, we sample sufficient configurations by random sampling from the entire space and local sampling on well-performed observed configurations~\cite{hutter2011sequential}. 
Then we calculate the performance and diversity acquisition value for each sampled configuration. 
Based on these values, we further rank the sampled configurations and obtain the ranking value of $x_i$ as $R_{perf}(x_i)$ and $R_{div}(x_i)$ for the performance and diversity acquisition function, respectively.
Finally, given a configuration $x_i$, we define the weighted acquisition function $\alpha$ for \sys as:
\begin{equation}
\label{eq:final_acq}
    \alpha(x_i)=R_{perf}(x_i)+wR_{div}(x_i),\ \quad w=\beta(sigmoid(\tau t)-0.5),
\end{equation}
where $w$ is the weight for the diversity acquisition, $t$ is the current number of BO iterations, $\beta$ and $\tau$ are two hyperparameters. 
To match the intuition that the optimization should focus on performance in the beginning and gradually shift its attention to diversity, \sys applies a saturating weight, where $w$ ranges from $[0,\beta)$ and $\tau$ controls the speed of approaching saturation.
The goal of \sys is to suggest configurations that \textbf{minimize} the function in Equation~\ref{eq:final_acq}. 

\begin{algorithm}[tpb]
  \caption{Algorithm framework of \sys.}
  \label{alg:framework}
    \KwIn{The search budget $\mathcal{B}$, the architecture search space $\mathcal{X}$, the ensemble size $E$, the training and validation set $\mathcal{D}_{train}$, $\mathcal{D}_{val}$.}
    Initialize observations as $D=\varnothing$\;
    \While{{budget $\mathcal{B}$ does not exhaust} }{
     \eIf{$|D|<5$}
     {Suggest a random configuration $\tilde{x} \in \mathcal{X}$\;}
     {
     Fit performance surrogate $M_{perf}$ and diversity surrogate $M_{div}$ based on observations $D$\;
     Build a temporary pool of configurations as  $\mathcal{P}=\{\theta_1,...,\theta_E\}=EnsembleSelection(D,\mathcal{D}_{val},E)$\;
     Compute the ranks of sampled configurations $R_{perf}$ and $R_{div}$ based on the performance and diversity surrogates $M_{perf}, M_{div}$ and the temporary pool $\mathcal{P}$\;
     Suggest a configuration  $\tilde{x}=\mathop{\arg\min}\limits_{x \in \mathcal{X}}\alpha(x)$ based on Equation~\ref{eq:final_acq}\;
     }
     Build and train the learner $f_{\tilde{x}}$ on $\mathcal{D}_{train}$ and evaluate its performance on $\mathcal{D}_{val}$ as $\tilde{y}$\;
     Update the observations $D=D \cup \{(\tilde{x},\tilde{y})\}$; 
     }
     
     Generate a pool of base configurations $\{\theta_1,...,\theta_E\}=EnsembleSelection(D,\mathcal{D}_{val},E)$;
     
    \textbf{return} the final ensemble $Ensemble(T_{\theta_1},...,T_{\theta_E})$.
\end{algorithm}

Algorithm~\ref{alg:framework} illustrates the procedure of \sys.
During each iteration after initialization, \sys 1) fits the performance and diversity surrogates based on observations (Line 6); 
2) builds a temporary configuration pool by applying ensemble selection on the observation history (Line 7); 
3) samples candidate configurations and compute their ranking values (Line 8);
4) suggests a configuration that minimizes the combined ranking value in Equation~\ref{eq:final_acq} (Line 9); 
5) evaluates the suggested configuration on the validation set and then updates the observations (Lines 10-11).

\subsection{Discussion}
\label{sec:discussion}
In this section, we provide the discussion on \sys as follows:


\textbf{Time complexity.}
As mentioned in Section~\ref{sec:surrogate}, the time complexity of fitting the diversity surrogate is $O(|D|^2\log|D|)$, and the complexity of building a temporary pool is $O(|\mathcal{D}_{val}||D|)$, where $|\mathcal{D}_{val}|$ is the number of validation data samples, and $|D|$ is the number of observations.
For large datasets, we can prepare a constant subset of validation samples for ensemble selection.
Therefore, the time complexity depends on the choice of performance surrogate in BO.
Concretely, the complexity of \sys for each iteration is $O(|D|^3)$ when using the Gaussian Process~\cite{snoek2012practical} and $O(|D|^2\log|D|)$ when using the probabilistic random forest~\cite{hutter2011sequential}.

\textbf{Extension.}
As an abstract algorithm framework, the components of \sys can be replaced to meet different requirements. 
Though this paper focuses on CASH problems for classification, \sys can also be applied to regression problems by defining a new diversity function based on regression predictions instead of Equation~\ref{eq:div}.
In addition, \sys is independent of the choice of performance surrogate for Bayesian optimization, i.e., the performance surrogate can be replaced with state-of-the-art ones proposed for specific scenarios.

\textbf{Foundation.}
Like the diversity-driven methods in other scenarios~\cite{zhang2020efficient,partalas2008focused}, \sys does not target at optimizing Equation~\ref{eq:cash_es} directly. 
The foundation of its effectiveness lies in the claim that it is beneficial for ensembles to not only have promising base learners, but also more diversity in their
predictions.
The claim has been studied by extensive theoretical work~\cite{lecun2015deep,bian2021does,hansen1990neural,zhou2002ensembling,kuncheva2003measures}, and please refer to previous work for more details.
In the following section, we will empirically show that the ensemble generated by \sys outperforms the state-of-the-art methods in real-world CASH problems.

\textbf{Limitations.}
The use of ensemble leads to higher inference latency than using the single best learner (approximately K times where K is the number of learners in the ensemble). This latency can be reduced with the aid of parallel computing if we have sufficient computational resources; As ensemble selection is fitted on the validation set, there's a risk of overfitting on the test set for small datasets; \sys using Equation 3 as the diversity function can not directly support algorithms that only predict class labels (e.g., SVC). Though \sys still works by converting the outputs to class probability (like [1, 0, ...]), other diversity functions can be developed to support those algorithms better.

\section{Experiments}
\label{sec:exp}
In this section, we evaluate our proposed method on real-world CASH problems using public datasets. 
In the following, we list three main insights that we will investigate: 
1) The diversity surrogate in \sys can predict the diversity value of two unseen configurations well.
2) The \sys framework outperforms the post-hoc designs used in recent AutoML systems and other competitive baselines for ensemble learning, in terms of both validation and test performance.
3) The base learners in the ensemble given by \sys show similar average performance but enjoy higher diversity than those from other post-hoc designs. 

\subsection{Experiment Setup}
\textbf{Baselines.}
We compare the proposed \sys with the following eight baselines 
--- 
\textit{Three CASH methods:}
1) Random search (RS)~\cite{bergstra2012random};
2) Bayesian optimization (BO)~\cite{hutter2011sequential};
3) Rising Bandit (RB)~\cite{li2020efficient};
---
\textit{Two AutoML methods proposed for ensemble learning:}
4) Ensemble optimization (EO)~\cite{levesque2016bayesian}; 5) Neural ensemble search (NES)~\cite{zaidi2021neural};
---
\textit{Three post-hoc designs:}
6) Random search with post-hoc ensemble (RS-ES);
7) Bayesian optimization with post-hoc ensemble (BO-ES)~\cite{feurer2015efficient}: the default strategy in Auto-sklearn;
8) Rising bandit with post-hoc ensemble (RB-ES)~\cite{li2021volcanoml}: the default strategy in VolcanoML.
While for \sys, we also implement the variant without post-hoc ensemble, which we denote as ``\sys-''.

\textbf{Datasets and search space.}
While recent AutoML systems differ in both search space and algorithm, to make a fair comparison of algorithms,
we conduct the experiments on the same search space. 
Concretely, we slightly modify the search space of the AutoML system VolcanoML~\cite{li2021volcanoml}.
The search space contains 100 hyperparameters in total, and the details of algorithms and feature engineering hyperparameters are provided in Appendix A.3.
In addition, we use 15 public classification datasets that are collected from OpenML~\cite{vanschoren2014openml}, whose number of samples ranges from 2k to 20k.
More details about the datasets are provided in Appendix A.2.

\textbf{Basic settings.}
Each dataset is split into three sets, which are the training (60\%), validation (20\%), and test (20\%) sets.
To evaluate the diversity surrogate, we use the Kendall-tau rank correlation as the metric since we only care about the ranking relationship between two pairs during optimization.
For comparison with other baselines on CASH problems, we report the best-observed validation error during optimization and the final test error.
While it takes a different amount of time to evaluate the same configuration on different datasets, we use the evaluation iterations as the unit of budget. Following VolcanoML~\cite{li2021volcanoml} where each baseline evaluates approximately 250 configurations, we set the number of maximum iterations to 250.
The evaluation of each method on each dataset is repeated 10 times, and we report the mean$\pm$std. result by default.
The hyperparameters $\beta$ and $\tau$ are set to 0.05 and 0.2 in \sys.
We provide more implementation details for other baselines and the sensitivity analysis in Appendix A.4 and A.5, respectively.


\subsection{Evaluation of Diversity Surrogate}
\label{sec:exp_surrogate}
\begin{figure*}[tp!]
\centering  
\subfigure[quake]{
\includegraphics[width=0.45\textwidth]{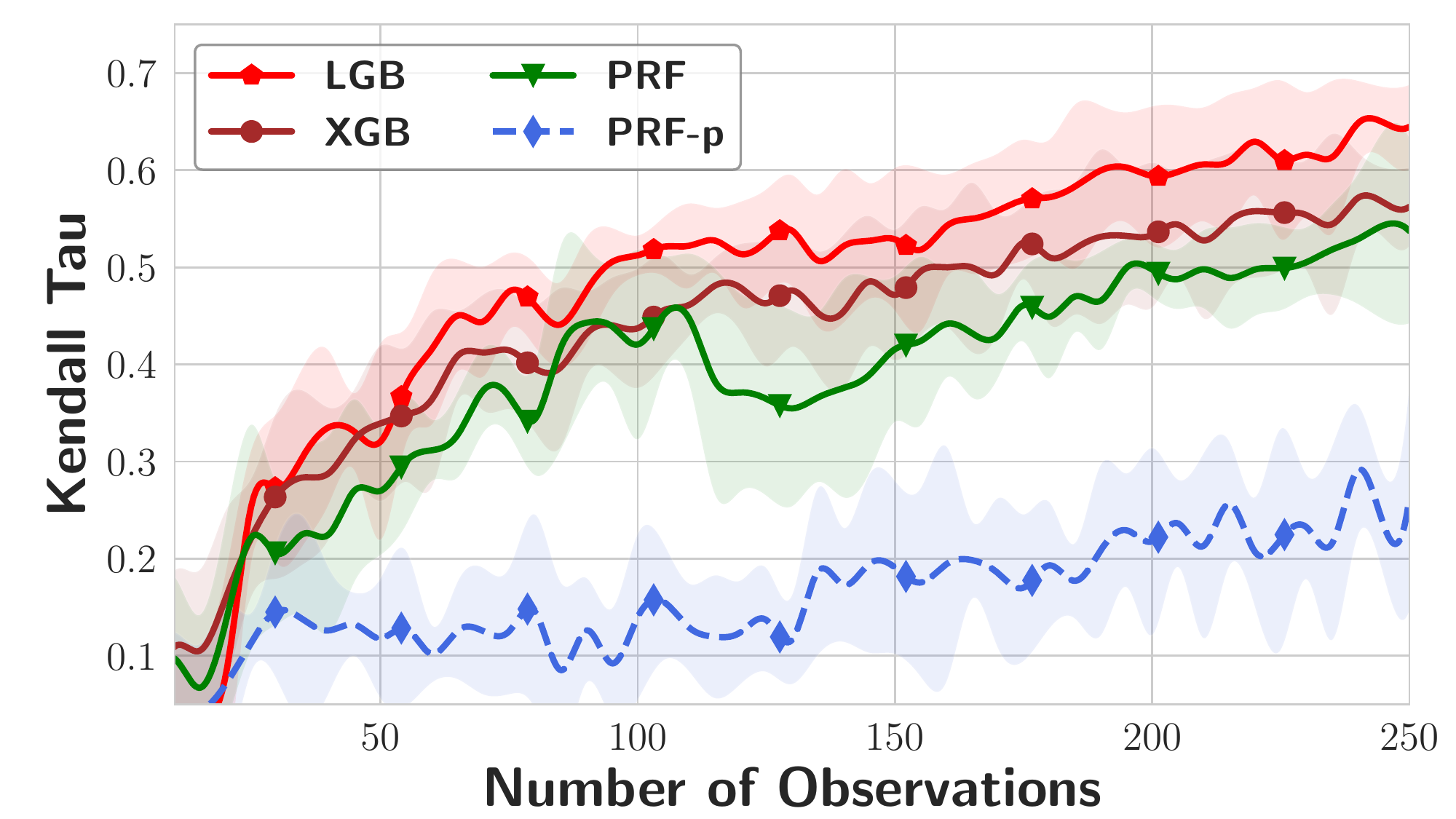}}\hspace{3mm}
\subfigure[wind]{
\includegraphics[width=0.45\textwidth]{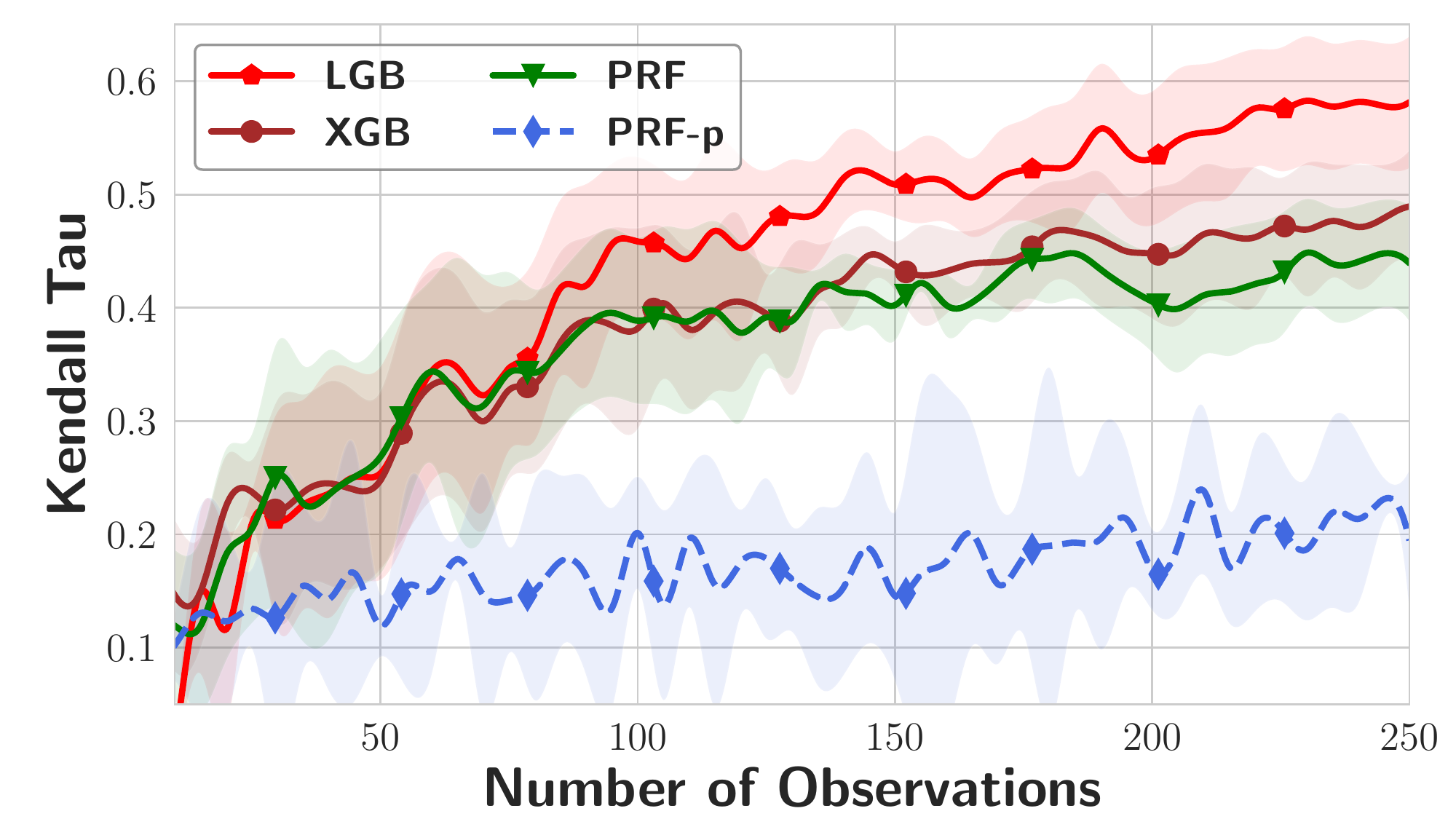}}
\caption{Kendall Tau correlation of different diversity surrogates with standard deviations shaded. `-p' refers to the performance surrogate while the other three are diversity surrogates.}
\label{fig:kt}
\end{figure*}

To show the soundness of \sys, we first provide an analysis of the fitting capability of different diversity surrogates during optimization.
We take LightGBM (LGB)~\cite{ke2017lightgbm}, XGBoost (XGB)~\cite{chen2016xgboost}, and probabilistic random forest (PRF) as the candidates for diversity surrogate, where all three tree-based alternatives share the same time complexity $O(|D|^2\log(|D|))$.
We evaluate 300 randomly chosen configurations, among which up to 250 configurations are used to fit the surrogate, and the left 50 are used for surrogate evaluation. 
Figure~\ref{fig:kt} shows the Kendall Tau correlation between the surrogate predictive means across different runs and the ground-truth results on two different datasets. 

\begin{figure*}[tp!]
\centering  
\includegraphics[width=0.97\textwidth]{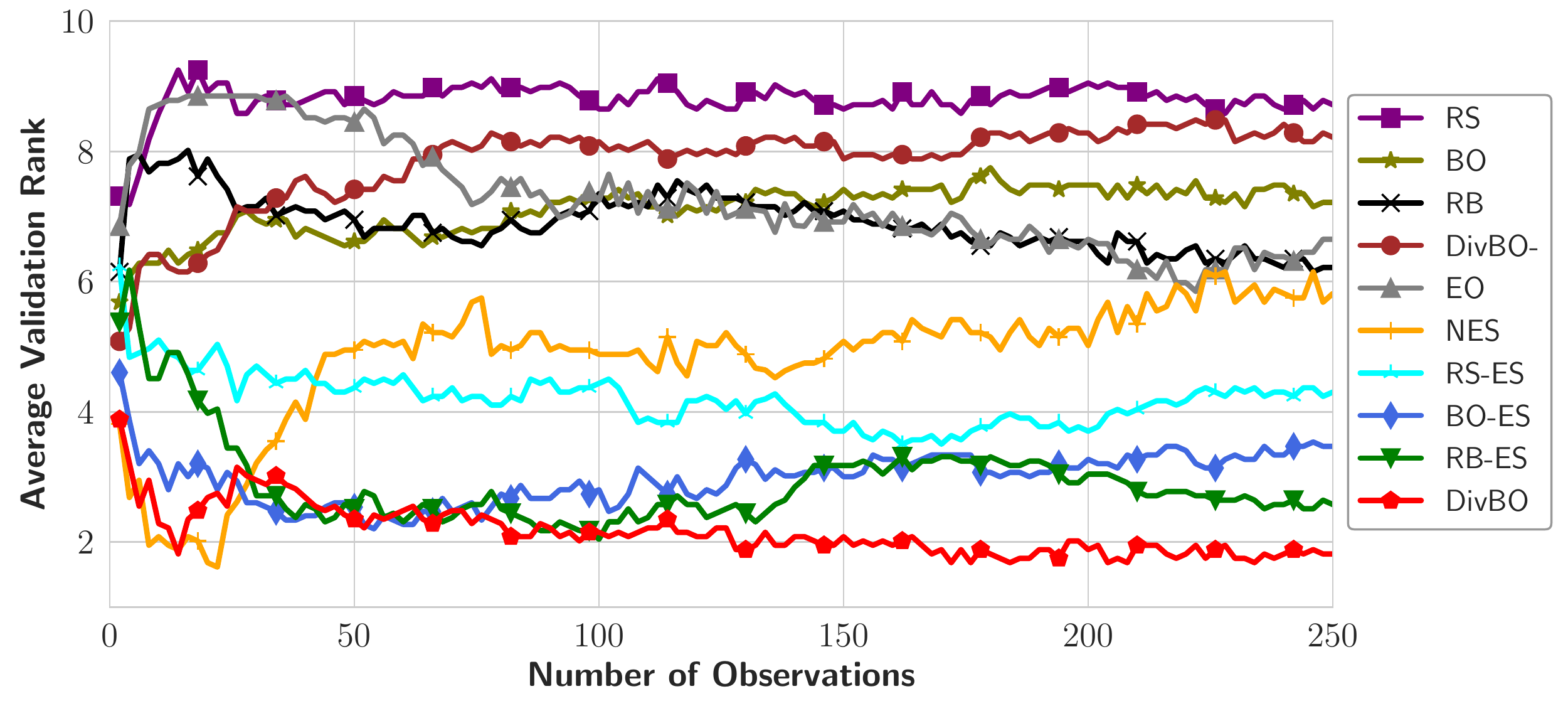}
\caption{Average validation rank of 10 methods during optimization across 15 datasets. Note that ranks are a relative measure of performance (the rank of all methods add up to 55), and an improvement in validation error in one method may influence the rank of another one.}
\label{fig:rank}
\end{figure*}

We observe that, in general, the correlation of the diversity surrogate improves as the observations increase. 
Among the three alternatives, XGBoost performs slightly better than PRF, and LightGBM performs better than the other two alternatives. 
Remarkably, LightGBM achieves a strong Kendall Tau correlation of 0.65 and 0.58 on quake and wind when fitted with 250 observations, respectively.
We also evaluate the performance surrogate (PRF-p) used in BO.
Its correlation over the number of observations is much lower than the diversity surrogate, and the correlation is only 0.26 and 0.19 on quake and wind when fitted with 250 observations.
The reason is that the search space for CASH problems is too large, and the performance surrogate can not be fitted well using limited observations $D$.
While the number of pair-wise samples is much larger than non-pair-wise samples (i.e., $|D|^2$ vs. $|D|$), the diversity surrogate captures more information from the observations.
As a result, the diversity surrogate can fit the diversity relationship between two configurations well and enjoys a relatively strong correlation given a limited budget.
In the following, we apply LightGBM as the diversity surrogate and evaluate \sys on real-world AutoML tasks. 

\subsection{Evaluation of \sys}
\label{sec:exp_divbo}

\begin{figure*}[tb]
\centering  
\subfigure[Pollen]{
\includegraphics[width=0.28\textwidth]{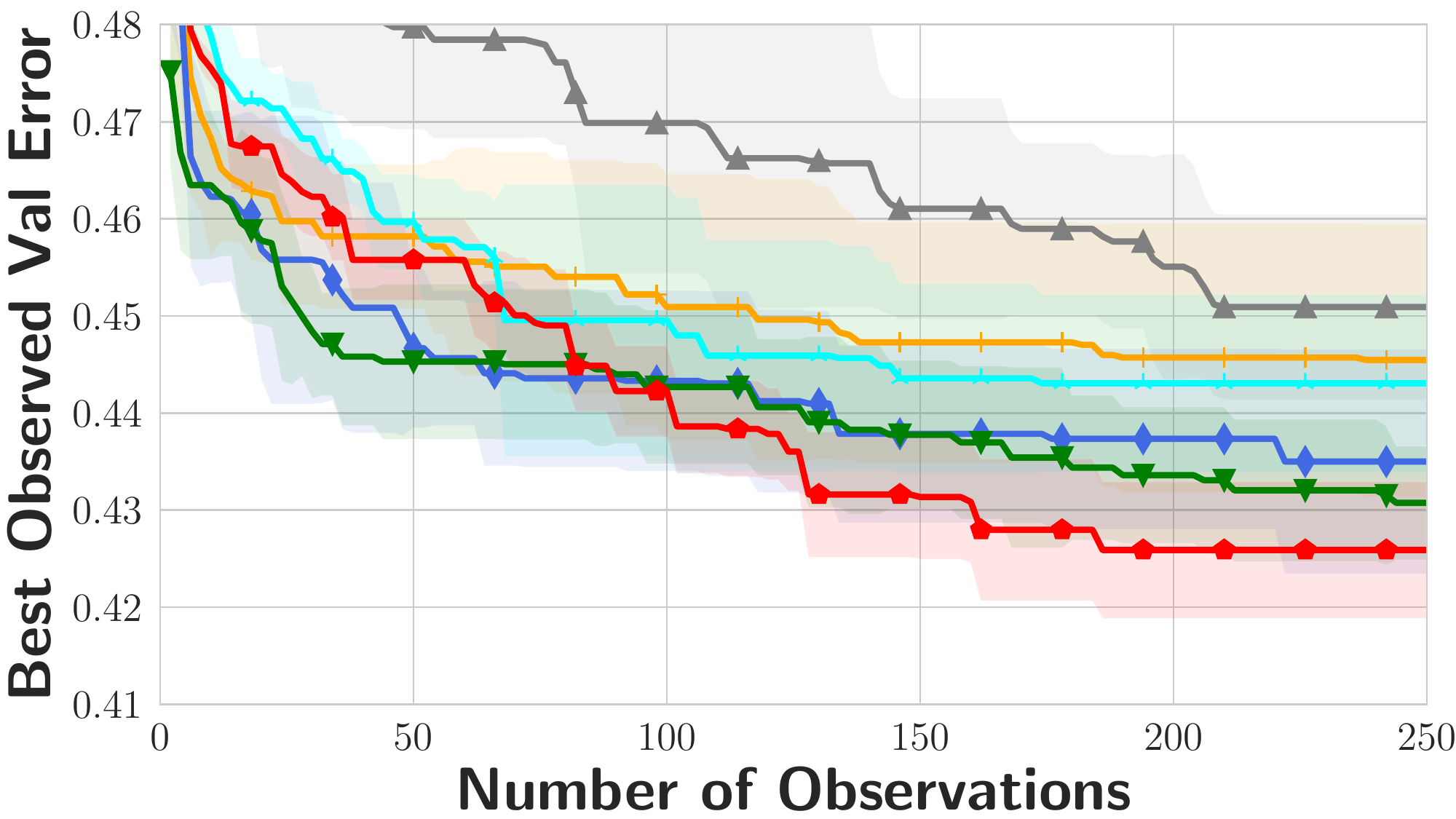}
\label{fig:val_pollen}
}
\subfigure[Wind]{
\includegraphics[width=0.28\textwidth]{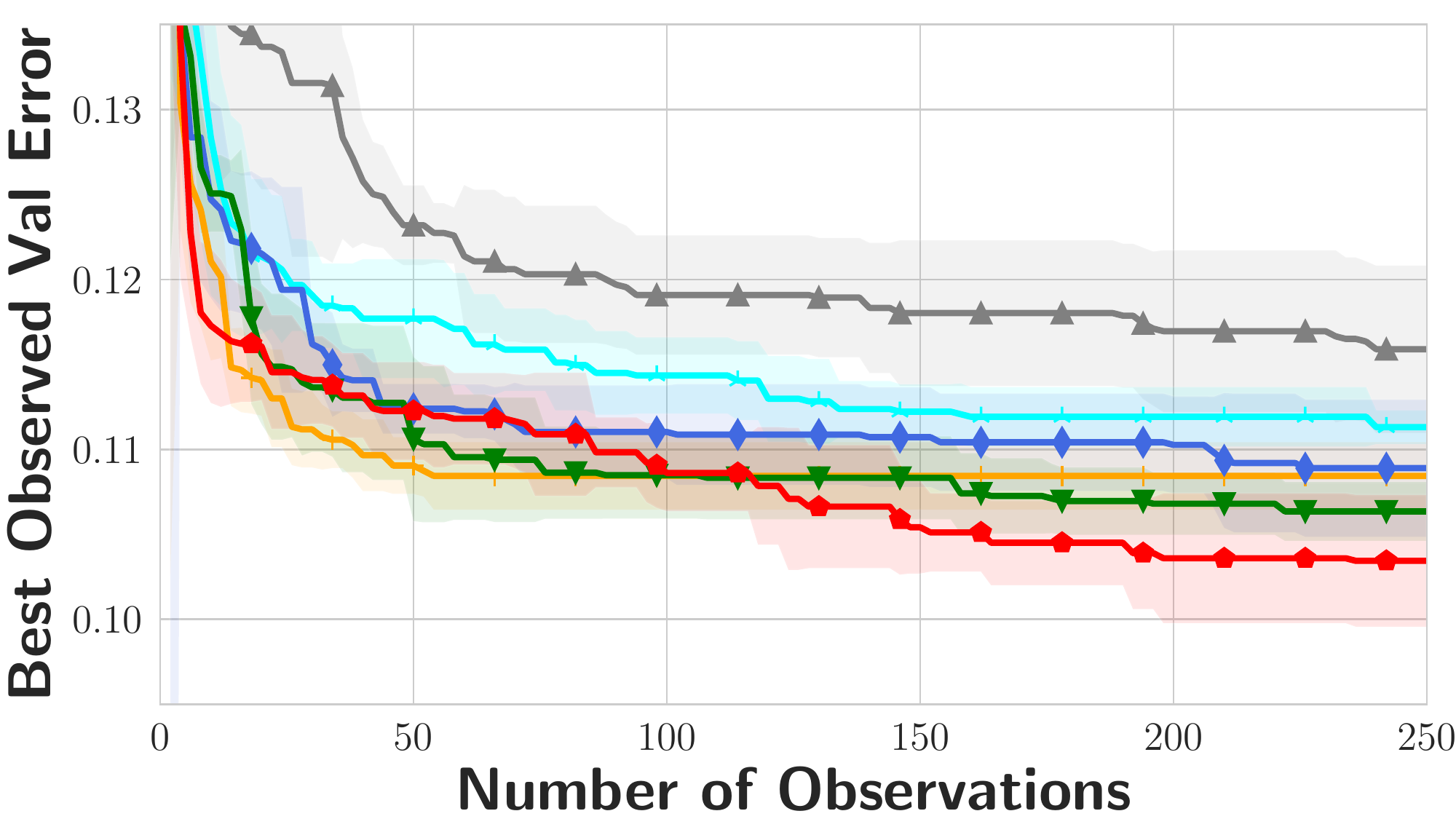}
\label{fig:val_wind}
}
\subfigure[Average test rank]{
\includegraphics[width=0.28\textwidth]{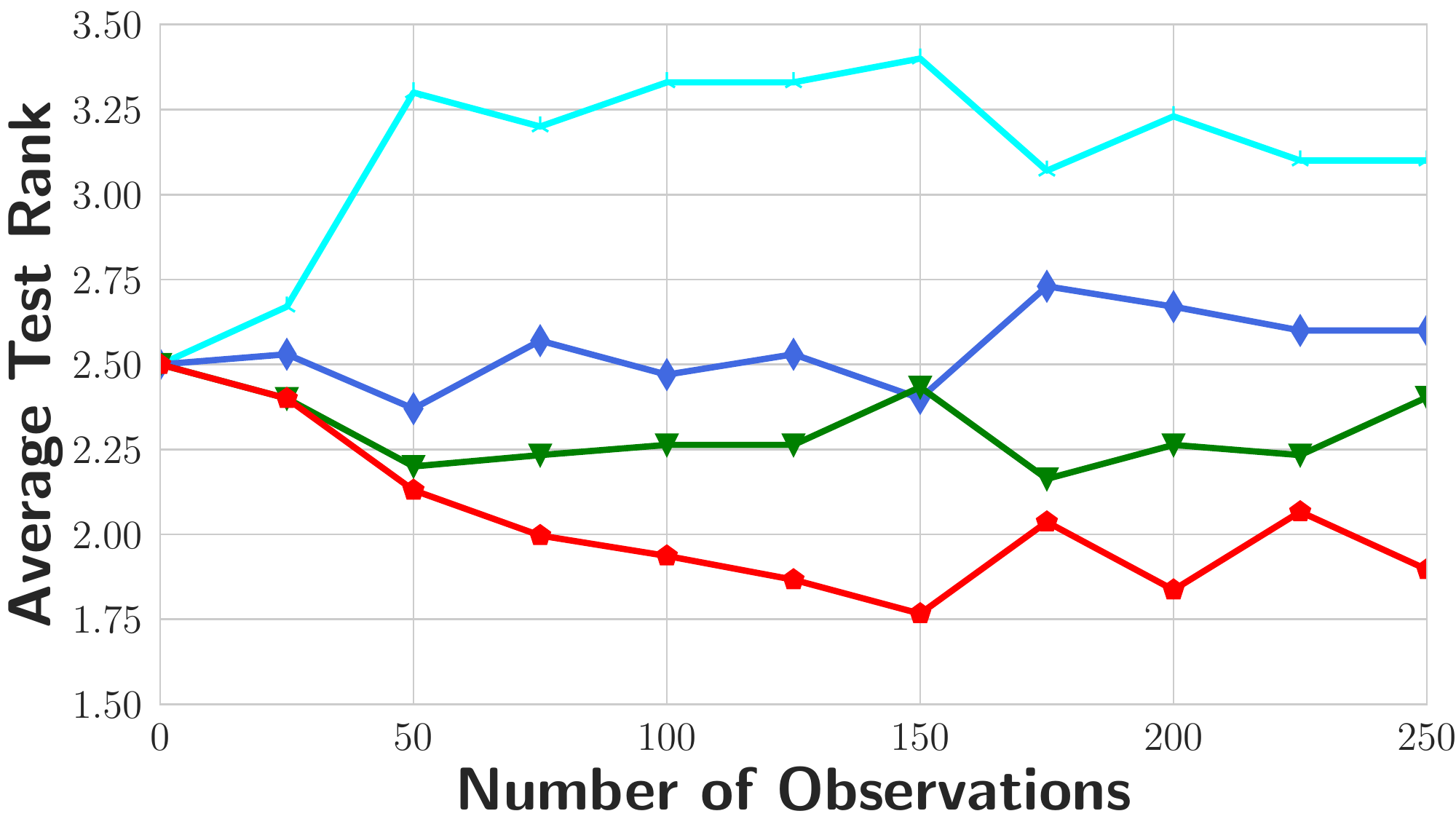}
\label{fig:test_rank}
}
\subfigure{
\raisebox{0.19\height}{\includegraphics[width=0.09\textwidth]{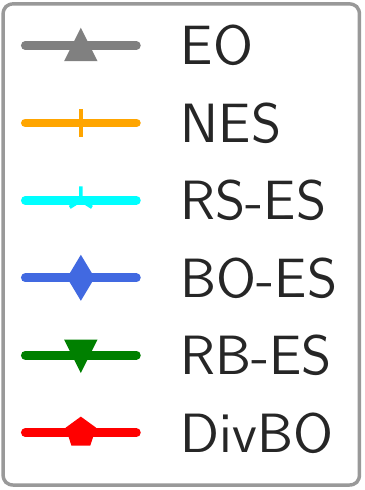}}}

\caption{Figures (a) and (b): Best observed validation error with standard deviations shaded on two datasets. Figure (c): Average test rank of post-hoc ensemble designs across 15 datasets.}
\label{fig:validation}

\end{figure*}

\textbf{Performance Analysis.}
In this section, we evaluate \sys on 15 real-world CASH problems. 
Figure~\ref{fig:rank} shows the rank of validation error of all compared methods during optimization.
We get four observations from the figure:
1) Ensemble learning indeed helps improve the performance of CASH results. Though Rising Bandit (RB) is competitive, the rank of non-ensemble methods (RS, BO, \sys-) at the 250$^{th}$ iteration is only 8.72, 7.21, and 8.22 among 10 compared methods, respectively;
2) Among methods with ensemble learning, the post-hoc designs outperform the other designs for ensemble learning (NES, EO). 
Note that, random search with ensemble selection (RS-ES) is a strong baseline.
It achieves a rank of 4.30 at the 250$^{th}$ iteration, which is better than that of NES (5.82) and EO (6.65).
The reason is that the evolutionary algorithm in NES is not suitable for large search space while the performance of EO fluctuates if a poor learner is added into its fixed ensemble during optimization;
3) Among post-hoc designs, \sys outperforms the other ones.
While the second-best baseline achieves a rank of 2.58 at the 250$^{th}$ iteration,
the rank of \sys is 1.82, and it consistently outperforms the other baselines after about 100 iterations.
4) We also find that \sys- performs worse than BO without ensemble learning.
The reason is that, the consideration of diversity is useless when searching for the single best classifier.
But with the aid of the performance surrogate, \sys- still performs better than random search.
To demonstrate the validation error on specific datasets, we also plot the best-observed validation errors during optimization for six methods with ensemble learning in Figures~\ref{fig:val_pollen} and~\ref{fig:val_wind}. 
Similar to the trend of rank, \sys exceeds the second-best baseline RB-ES 
after about 100 iterations.
We further compare the number of required iterations for \sys to achieve the same best validation errors as other methods.
Concretely, \sys achieves 1.52-1.54x and 1.67-2.53x speedups relative to RB-ES and BO-ES on the two datasets.
We provide ablation study on weight scheduling and comparison with other intuitive designs in Appendix A.5. 

Table~\ref{tab:test} demonstrates the test errors and average rank on 15 datasets.
We observe that the rank value of \sys on test error is sometimes not consistent with that on validation error.
The reason is that the distributions of the validation and test set are not exactly the same~\cite{hutter2019automated}.
In other words, the configuration with the best validation error may not be the one with the best test error.
But overall, post-hoc ensemble designs outperform other methods for ensemble learning, among which \sys achieves the best test error on 10 out of 15 datasets, and its average rank is 1.73.
The second best baseline is RB-ES, which performs the best on 3 out of 15 datasets, and its average rank is 2.90.
The test rank of 4 post-hoc ensemble designs during optimization is also presented in Figure~\ref{fig:test_rank}.

To check whether the improvement of \sys is statistically significant, we conduct the Wilcoxon signed-rank test on each dataset given two methods. The difference is significant when the value $p \leq 0.05$. 
We count the number of datasets if 1) the mean error of \sys is lower, and the difference is statistically significant (B); 2) the difference is not statistically significant (S); and 3) the mean error of \sys is higher, and the difference is statistically significant (W). 
The results are presented in Table~\ref{tab:significance}.
Though RB-ES is a strong baseline, we observe that \sys performs no worse than RB-ES on 12 datasets and better on 8 datasets. In Appendix A.5, we provide additional experiments to study the effects of different ensemble strategies and ensemble sizes. 



\begin{table*}[t]
\caption{Test error (\%) with standard deviations and the average rank across different datasets.}
\centering
\noindent
\resizebox{0.99\linewidth}{!}{
\subtable{
\begin{tabular}{ccccccccc}
\toprule
Method&amazon\_employee&bank32nh&cpu\_act&cpu\_small&eeg&elevators&house\_8L&pol\\
\midrule
\multicolumn{9}{c}{\textbf{CASH Methods}}\\
RS & $5.32\pm0.08$ & $18.43\pm0.79$& $6.17\pm0.45$& $7.57\pm0.41$& $6.58\pm0.52$& $10.22\pm0.68$& $11.32\pm0.26$& $1.65\pm0.12$ \\
BO & $5.26\pm0.06$ & $18.46\pm0.50$& $5.69\pm0.38$& $7.76\pm0.92$& $5.47\pm1.35$& $10.25\pm0.87$& $11.20\pm0.19$& $1.59\pm0.50$ \\
RB & $5.27\pm0.08$ & $18.33\pm0.32$& $5.70\pm0.31$& $7.53\pm0.35$& $4.70\pm1.05$& $9.77\pm0.20$& $11.12\pm0.15$& $1.39\pm0.03$ \\
\midrule
\multicolumn{9}{c}{\textbf{Methods for Ensemble Learning}}\\
EO & $5.19\pm0.27$ & $18.47\pm0.53$& $5.85\pm0.62$& $7.42\pm0.45$& $3.54\pm0.76$& $10.34\pm0.61$& $11.14\pm0.21$& $1.43\pm0.13$ \\
NES & $5.32\pm0.38$ & $18.21\pm0.32$& $5.59\pm0.20$& $7.28\pm0.84$& $\bm{2.68\pm0.73}$& $9.75\pm0.30$& $11.28\pm0.48$& $1.74\pm0.35$ \\
\midrule
\multicolumn{9}{c}{\textbf{Post-hoc Designs}}\\
RS-ES & $5.29\pm0.15$ & $18.30\pm0.66$& $5.70\pm0.26$& $7.21\pm0.30$& $4.45\pm0.22$& $9.51\pm0.28$& $11.21\pm0.38$& $1.39\pm0.15$ \\
BO-ES & $5.25\pm0.15$ & $18.41\pm0.39$& $5.50\pm0.47$& $7.16\pm0.28$& $3.55\pm0.78$& $9.61\pm0.36$& $11.06\pm0.33$& $1.35\pm0.18$ \\
RB-ES & $5.21\pm0.11$ & $\bm{18.07\pm0.57}$& $5.58\pm0.20$& $7.08\pm0.22$& $2.86\pm0.92$& $10.01\pm0.15$& $10.81\pm0.27$& $1.36\pm0.16$ \\
\sys & $\bm{5.16\pm0.09}$ & $18.35\pm0.32$& $\bm{5.37\pm0.23}$& $\bm{7.04\pm0.29}$& $3.26\pm0.84$& $\bm{9.40\pm0.28}$& $\bm{10.80\pm0.22}$& $\bm{1.34\pm0.17}$ \\
\bottomrule
\end{tabular}
}}
\\
\resizebox{0.99\linewidth}{!}{
\subtable{
\begin{tabular}{ccccccccc}
\toprule
Method & pollen & puma32H & quake & satimage & spambase & wind & 2dplanes & Average Rank\\
\midrule
\multicolumn{9}{c}{\textbf{CASH Methods}}\\
RS & $51.53\pm0.53$ & $10.59\pm1.26$& $48.17\pm1.88$& $10.00\pm0.72$& $7.14\pm1.23$& $14.66\pm0.58$ & $7.21\pm0.08$& $8.37$ \\
BO & $49.64\pm3.39$ & $10.42\pm0.81$& $46.88\pm2.16$& $9.21\pm0.99$& $6.45\pm0.84$& $14.52\pm0.60$& $7.15\pm0.06$& $6.47$ \\
RB & $49.79\pm1.15$ & $11.21\pm0.39$& $47.98\pm1.56$& $9.50\pm0.94$& $6.71\pm1.05$& $14.11\pm0.25$& $7.20\pm0.05$& $6.37$ \\
\midrule
\multicolumn{9}{c}{\textbf{Methods for Ensemble Learning}}\\
EO & $49.01\pm2.10$ & $9.74\pm1.60$& $46.88\pm1.48$& $9.42\pm1.11$& $6.41\pm0.68$& $14.62\pm0.48$& $7.13\pm0.07$& $5.77$ \\
NES & $51.56\pm1.53$ & $10.63\pm0.54$& $46.42\pm1.15$& $8.66\pm0.95$& $6.23\pm0.68$& $14.25\pm0.50$& $7.11\pm0.08$& $5.00$ \\
\midrule
\multicolumn{9}{c}{\textbf{Post-hoc Designs}}\\
RS-ES & $49.69\pm1.63$ & $10.58\pm0.73$& $46.79\pm1.57$& $9.35\pm0.73$& $6.45\pm0.23$& $14.34\pm0.47$& $7.11\pm0.12$ & $5.27$ \\
BO-ES & $\bm{48.91\pm1.75}$ & $9.27\pm1.13$& $46.10\pm2.52$& $9.10\pm0.87$& $6.38\pm0.64$& $14.04\pm0.53$& $7.07\pm0.08$& $3.13$ \\
RB-ES & $49.58 \pm1.34$ & $\bm{7.85\pm0.43}$& $46.70\pm1.34$& $\bm{8.55\pm1.36}$& $6.12\pm0.36$& $13.98\pm0.45$&$7.20\pm0.08$ & $2.90$ \\
\sys & 
$49.25\pm1.35$ & $8.07\pm0.99$& $\bm{45.55\pm1.37}$& $8.71\pm1.25$& $\bm{5.91\pm0.45}$& $\bm{13.93\pm0.42}$& $\bm{7.00\pm0.08}$& $\bm{1.73}$ \\
\bottomrule
\end{tabular}
}}

\label{tab:test}
\end{table*}

\begin{table}[t]
    \centering
    \caption{Counts of datasets when \sys performs statistically better (B), the same (S), and worse (W) than compared three baselines.}

    \resizebox{0.25\linewidth}{!}{
    \subtable[vs. RS-ES]{
    \begin{tabular}{cccc}
    \toprule
         & B & S & W  \\
    \hline
    \sys & 13 & 2 & 0 \\
    \bottomrule
    \end{tabular}
    }}
    \quad 
    \resizebox{0.25\linewidth}{!}{
    \subtable[vs. BO-ES]{
    \begin{tabular}{cccc}
    \toprule
         & B & S & W  \\
    \hline
    \sys & 12 & 1 & 2 \\
    \bottomrule
    \end{tabular}
    }}
    \quad 
    \resizebox{0.25\linewidth}{!}{
    \subtable[vs. RB-ES]{
    \begin{tabular}{cccc}
    \toprule
         & B & S & W  \\
    \hline
    \sys & 8 & 4 & 3 \\
    \bottomrule
    \end{tabular}
    }}
    \label{tab:significance}
\end{table}

\textbf{Diversity Analysis.}
Finally, we analyze the optimization process of BO-ES, RB-ES, and \sys.
In Table~\ref{tab:single_errors}, we show the validation errors of learners during optimization. 
Without ensemble, the single learner suggested by \sys- performs worse than BO and RB. Note that, this does not mean that \sys suggests bad configurations. We randomly evaluate 300 configurations from the search space. The mean result of those diverse configurations is better than 88\% of the random configurations. 
To show how the diversity during the search process affects the ensemble, we use the update times of the temporary pool as a metric and the results are shown in Table~\ref{tab:update_counts}.
Since the temporary pool is built in the same way as the final ensemble, if the temporary pool changes, the configuration suggested at the previous iteration is included in the pool. 
Therefore, a change of the temporary pool at least indicates the suggested configuration affects the current ensemble. 
However, though the pool changes, the performance may not be improved due to the greedy mechanism, and thus we count the effective update times (i.e., the pool changes and the validation error of the ensemble decreases).
As the pool updates very frequently in the beginning, we only calculate the mean effective update times of \sys, RB-ES, and BO-ES on all datasets during the last 50 and 100 iterations. The pool is relatively stable in the last 50 iterations, which also indicates a budget of 250 iterations is sufficient for the datasets. We observe that, on average, \sys will improve the temporary pool at least once in the last 50 iterations. 
While the difference between BO-ES and \sys is the diversity part, we attribute this frequency gain to the use of diversity during the search process.

In Figure~\ref{fig:exp_div}, given the total budget of 250 evaluation iterations, we plot the minimum diversity of the suggested learners (solid lines) after 200 iterations. 
The minimum diversity is defined as the diversity value (Equation~\ref{eq:div}) with the most similar learner in the ensemble built at the \textbf{previous} iteration.
The minimum diversity of suggestions given by BO-ES and RB-ES is around 0.05, which indicates that the configuration suggestion is similar to one of the configurations in the previous ensemble. 
While for \sys, the diversity is around 0.23, which is much higher than those of BO-ES and RB-ES.

\begin{minipage}[tb]{.51\textwidth}
  \begin{minipage}[thb]{\textwidth}
  \centering
  \captionof{table}{Val errors (\%) of single learners during optimization.}
  \resizebox{0.95\linewidth}{!}{
  \begin{tabular}{cccc}
    \toprule
    & BO-ES & RB-ES & \sys \\ 
    \midrule
    Val Errors (\%) & $44.03\pm2.58$ & $43.85\pm2.49$ & $44.33\pm2.67$ \\
    \bottomrule
\end{tabular}}
  \label{tab:single_errors}
\end{minipage}\\\\
\begin{minipage}[thb]{\textwidth}
  \centering
  \captionof{table}{Effective pool updates during optimization.}
  \resizebox{0.88\linewidth}{!}{
  \begin{tabular}{cccc}
    \toprule
    & BO-ES & RB-ES & \sys \\ 
    \midrule
    Counts (last 100) & 1.8 & 2.1 & 3.4 \\
    Counts (last 50) & 0.6 & 0.8 & 1.5 \\
    \bottomrule
\end{tabular}}
  \label{tab:update_counts}
\end{minipage}
\end{minipage}\quad 
\begin{minipage}[thb]{.45\textwidth}
  \centering
  \begin{center}
  \includegraphics[width=1\textwidth]{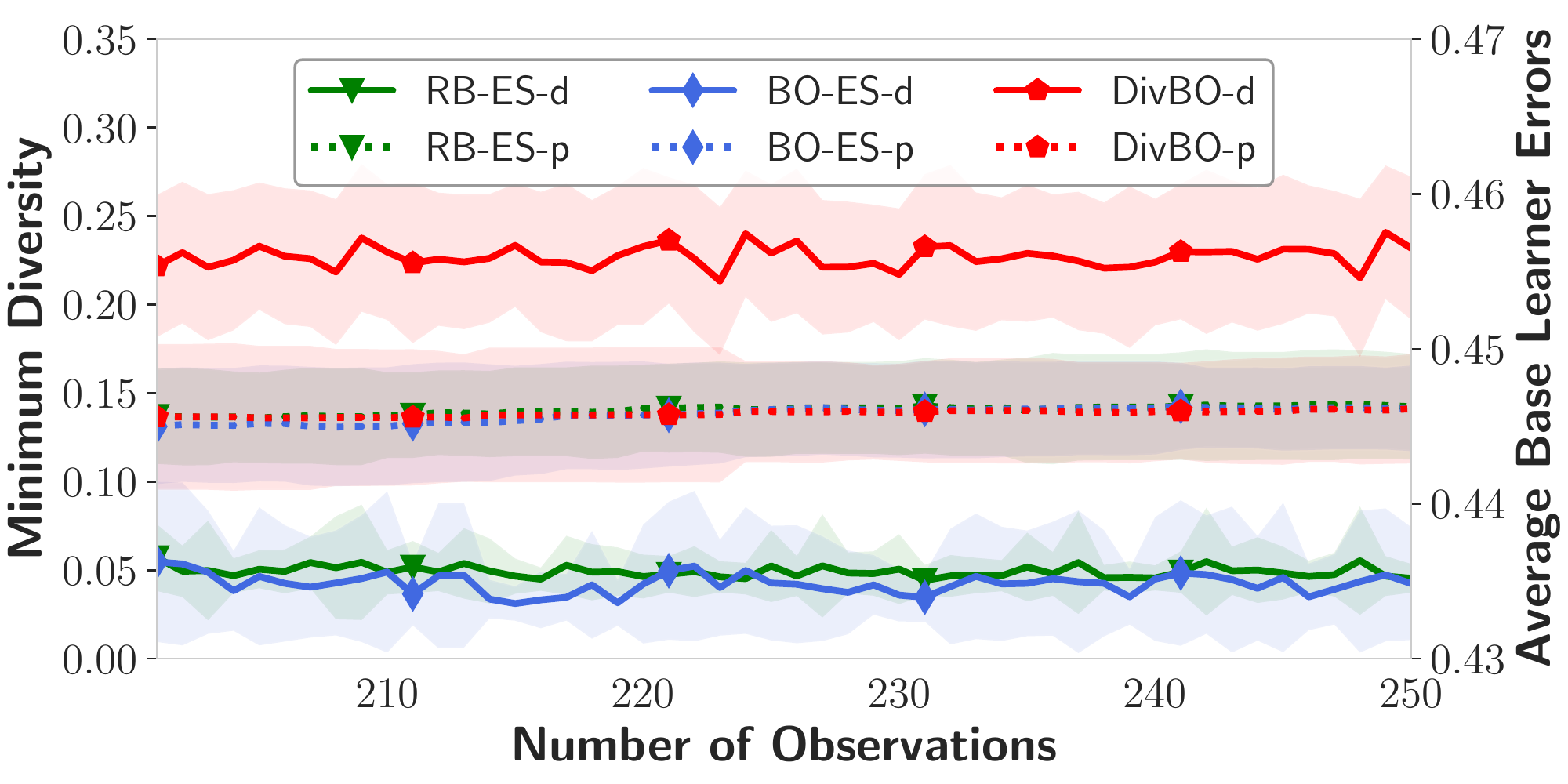}

  \end{center}
  \captionof{figure}{Minimum diversity of suggested learners (solid) and average error (dash) of base learners in the ensemble on quake.}
\label{fig:exp_div}
\end{minipage}

\begin{figure*}[t]
\centering  
\subfigure[RB-ES Disagreement: 0.13]{
\includegraphics[width=0.305\textwidth]{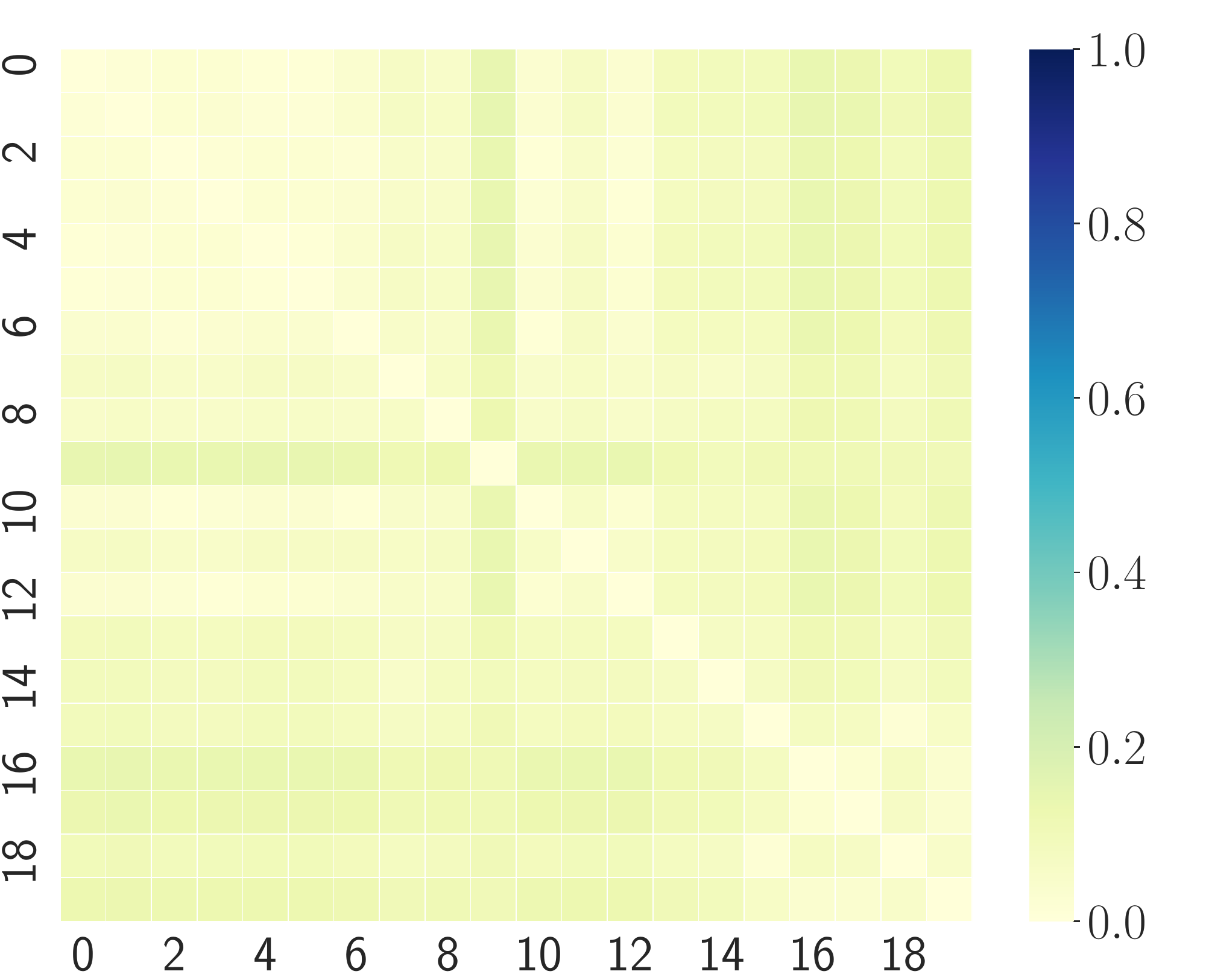}
}
\subfigure[BO-ES Disagreement: 0.09]{
\includegraphics[width=0.305\textwidth]{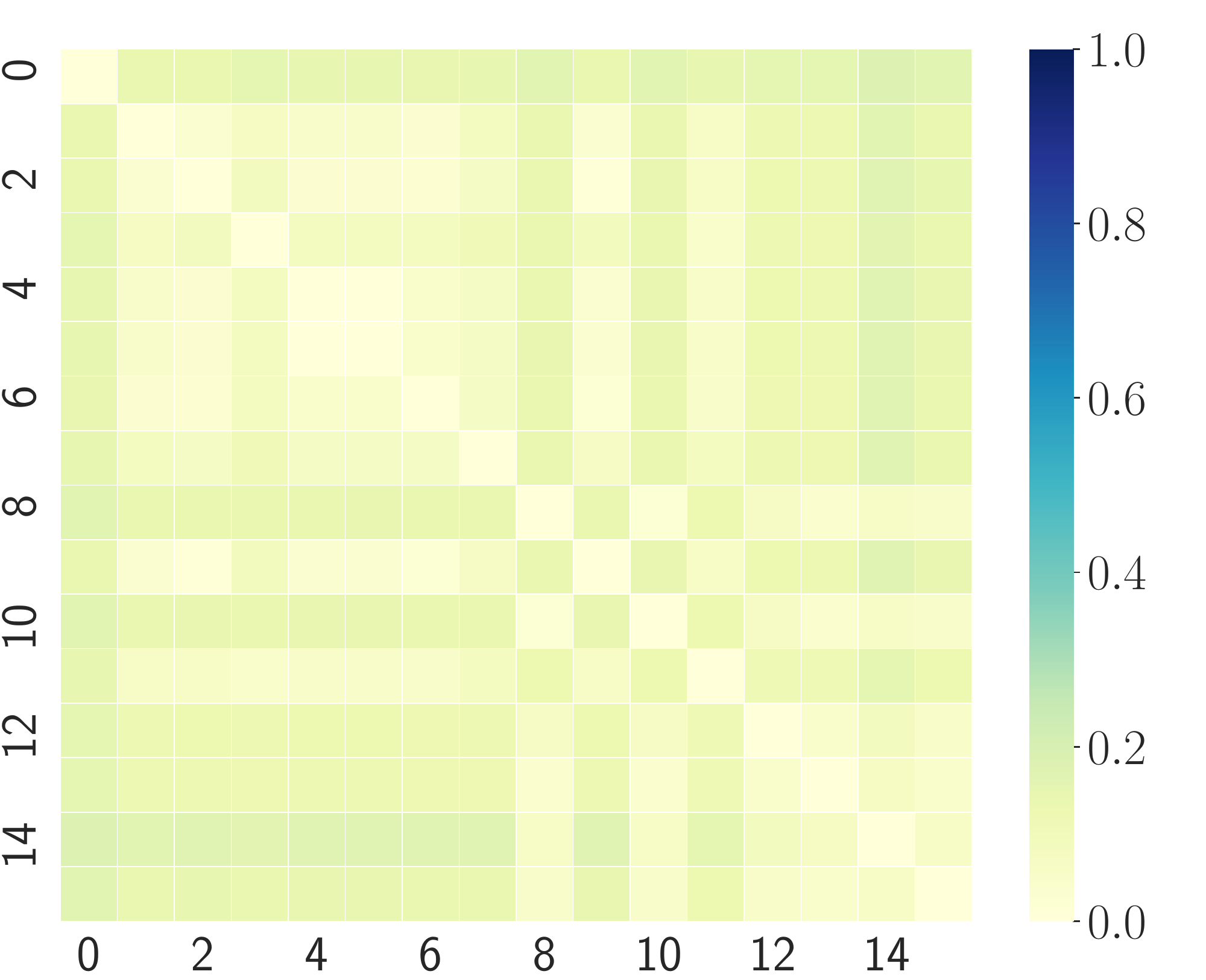}}
\subfigure[\sys Disagreement: 0.27]{
\includegraphics[width=0.305\textwidth]{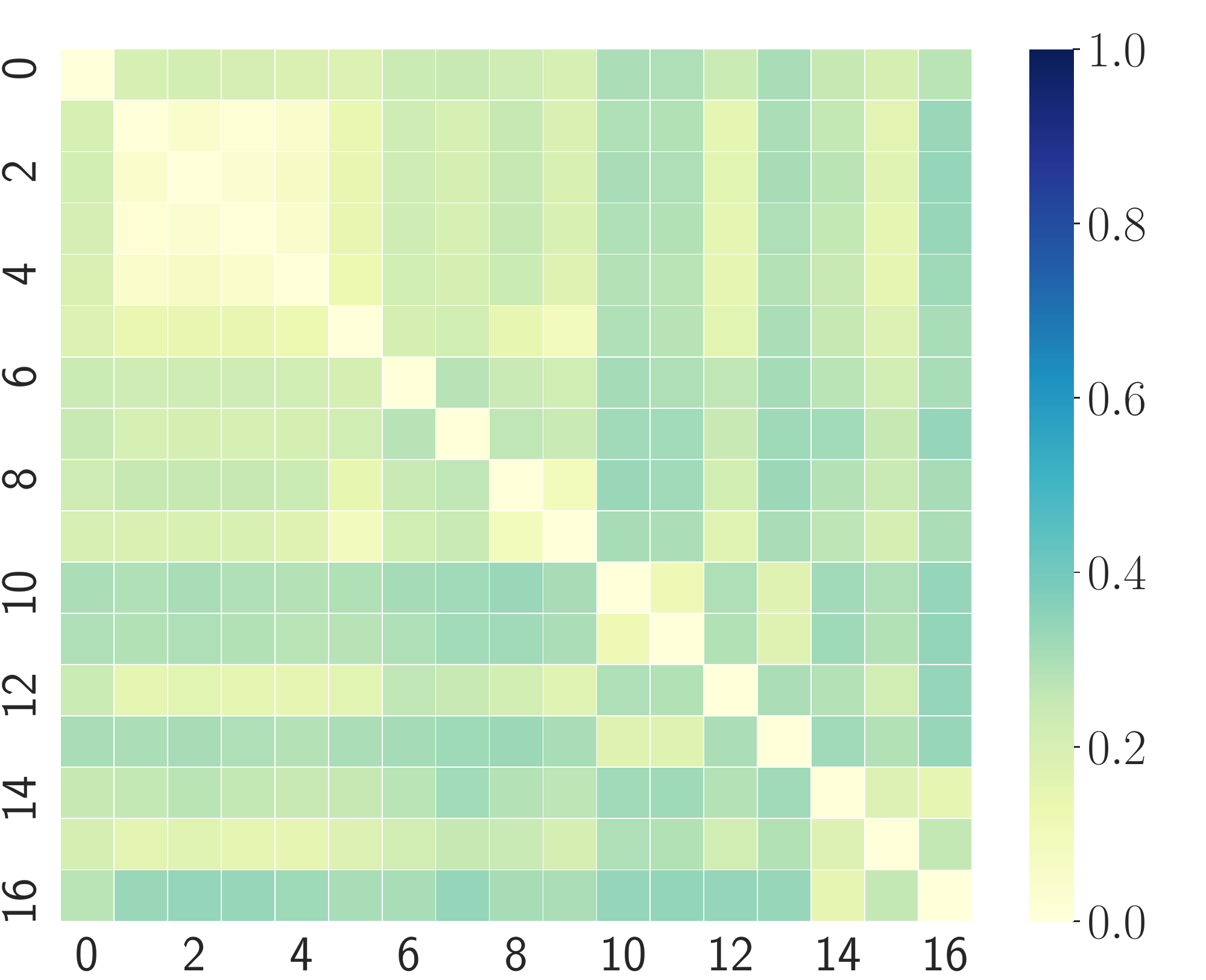}}
\caption{Pair-wise diversity of learners in the post-hoc ensemble on quake.
The learners in each method are numbered in order of observation.}
\label{fig:heatmap}
\end{figure*}

In addition, we plot the average validation error of base learners in the current ensemble (dash lines) in Figure~\ref{fig:exp_div}. 
The average error at the $250^{th}$ iteration given by the three post-hoc methods are quite similar (around $44.60\%$), which shows that applying the diversity surrogate in \sys will not degenerate the pool of base learners. 
We further plot the pair-wise diversity between unique learners in the final ensemble in Figure~\ref{fig:heatmap} and present the average pair-wise predictive disagreement~\cite{tang2006analysis} in the caption.
Generally, the larger disagreement is, the more diverse the learners in the ensemble are.
Since ensemble selection selects base learners with replacement, the number of unique learners in the ensemble is different in each independent run. 
Though the average performance of base learners in the three post-hoc methods are similar, for RB-ES and BO-ES, the base learners are quite similar to each other.
While for \sys, we find that the learners are more diverse.
Specifically, the learners that are found later (with a larger number) are more diverse with each other.
The reason is that the diversity surrogate becomes more precise when fitted with more observations, and thus it can suggest more diverse configurations.
This observation indicates that \sys is able to generate a more diverse ensemble while ensuring the performance of base learners.

\section{Conclusion}
\label{Conclusion}
In this paper, we introduced \sys, a diversity-aware framework based on Bayesian optimization to solve CASH problems for ensemble learning.
In \sys, we proposed to use a diversity surrogate to model the relationship between two configurations, and combined the ranking values of the performance and diversity surrogates with a saturating weight.
Through empirical study, we showed that the prediction of the diversity surrogate achieves a satisfactory correlation with ground-truth results, and \sys outperforms post-hoc designs in recent AutoML systems and other baselines for ensemble learning in CASH problems.

\section*{Acknowledgments}
This work is supported by NSFC (No. 61832001) and ZTE-PKU Joint Laboratory for Foundation Software. Yang Li and Bin Cui are the corresponding authors.


\nocite{liu2020value,zhu2021pre,zhang2020snapshot,li2021mfes,zhang2022pasca,yang2022diffusion,fan2021robust,limin2021balanced}

\bibliographystyle{abbrv}
\bibliography{reference}

\begin{thebibliography}{10}

\bibitem{bergstra2012random}
J.~Bergstra and Y.~Bengio.
\newblock Random search for hyper-parameter optimization.
\newblock {\em Journal of Machine Learning Research}, 13(Feb):281--305, 2012.

\bibitem{bergstra2011algorithms}
J.~S. Bergstra, R.~Bardenet, Y.~Bengio, and B.~K{\'e}gl.
\newblock Algorithms for hyper-parameter optimization.
\newblock In {\em Advances in neural information processing systems}, pages
  2546--2554, 2011.

\bibitem{bian2021does}
Y.~Bian and H.~Chen.
\newblock When does diversity help generalization in classification ensembles?
\newblock {\em IEEE Transactions on Cybernetics}, 2021.

\bibitem{bojer2021kaggle}
C.~S. Bojer and J.~P. Meldgaard.
\newblock Kaggle forecasting competitions: An overlooked learning opportunity.
\newblock {\em International Journal of Forecasting}, 37(2):587--603, 2021.

\bibitem{boyd2011distributed}
S.~Boyd, N.~Parikh, E.~Chu, B.~Peleato, J.~Eckstein, et~al.
\newblock Distributed optimization and statistical learning via the alternating
  direction method of multipliers.
\newblock {\em Foundations and Trends{\textregistered} in Machine learning},
  3(1):1--122, 2011.

\bibitem{breiman1996stacked}
L.~Breiman.
\newblock Stacked regressions.
\newblock {\em Machine learning}, 24(1):49--64, 1996.

\bibitem{caruana2004ensemble}
R.~Caruana, A.~Niculescu-Mizil, G.~Crew, and A.~Ksikes.
\newblock Ensemble selection from libraries of models.
\newblock In {\em Proceedings of the twenty-first international conference on
  Machine learning}, page~18, 2004.

\bibitem{chen2016xgboost}
T.~Chen and C.~Guestrin.
\newblock Xgboost: A scalable tree boosting system.
\newblock In {\em Proceedings of the 22nd acm sigkdd international conference
  on knowledge discovery and data mining}, pages 785--794. ACM, 2016.

\bibitem{dietterich2000ensemble}
T.~G. Dietterich.
\newblock Ensemble methods in machine learning.
\newblock In {\em International workshop on multiple classifier systems}, pages
  1--15. Springer, 2000.

\bibitem{fan2021robust}
G.~FAN, B.~LI, Q.~HAN, R.~JIAO, and G.~QU.
\newblock Robust lane detection and tracking based on machine vision.
\newblock {\em ZTE Communications}, 18(4):69--77, 2021.

\bibitem{feurer2015efficient}
M.~Feurer, A.~Klein, K.~Eggensperger, J.~Springenberg, M.~Blum, and F.~Hutter.
\newblock Efficient and robust automated machine learning.
\newblock In {\em Advances in neural information processing systems}, pages
  2962--2970, 2015.

\bibitem{goodfellow2014generative}
I.~Goodfellow, J.~Pouget-Abadie, M.~Mirza, B.~Xu, D.~Warde-Farley, S.~Ozair,
  A.~Courville, and Y.~Bengio.
\newblock Generative adversarial nets.
\newblock {\em Advances in neural information processing systems}, 27, 2014.

\bibitem{hansen1990neural}
L.~K. Hansen and P.~Salamon.
\newblock Neural network ensembles.
\newblock {\em IEEE transactions on pattern analysis and machine intelligence},
  12(10):993--1001, 1990.

\bibitem{he2016deep}
K.~He, X.~Zhang, S.~Ren, and J.~Sun.
\newblock Deep residual learning for image recognition.
\newblock In {\em Proceedings of the IEEE conference on computer vision and
  pattern recognition}, pages 770--778, 2016.

\bibitem{he2021automl}
X.~He, K.~Zhao, and X.~Chu.
\newblock Automl: A survey of the state-of-the-art.
\newblock {\em Knowledge-Based Systems}, 212:106622, 2021.

\bibitem{hoch2015ensemble}
T.~Hoch.
\newblock An ensemble learning approach for the kaggle taxi travel time
  prediction challenge.
\newblock In {\em DC@ PKDD/ECML}, 2015.

\bibitem{hutter2011sequential}
F.~Hutter, H.~H. Hoos, and K.~Leyton-Brown.
\newblock Sequential model-based optimization for general algorithm
  configuration.
\newblock In {\em International Conference on Learning and Intelligent
  Optimization}, pages 507--523. Springer, 2011.

\bibitem{hutter2019automated}
F.~Hutter, L.~Kotthoff, and J.~Vanschoren.
\newblock {\em Automated machine learning: methods, systems, challenges}.
\newblock Springer Nature, 2019.

\bibitem{ke2017lightgbm}
G.~Ke, Q.~Meng, T.~Finley, T.~Wang, W.~Chen, W.~Ma, Q.~Ye, and T.-Y. Liu.
\newblock Lightgbm: A highly efficient gradient boosting decision tree.
\newblock {\em Advances in neural information processing systems}, 30, 2017.

\bibitem{kuncheva2003measures}
L.~I. Kuncheva and C.~J. Whitaker.
\newblock Measures of diversity in classifier ensembles and their relationship
  with the ensemble accuracy.
\newblock {\em Machine learning}, 51(2):181--207, 2003.

\bibitem{lecun2015deep}
Y.~LeCun, Y.~Bengio, and G.~Hinton.
\newblock Deep learning.
\newblock {\em nature}, 521(7553):436--444, 2015.

\bibitem{levesque2016bayesian}
J.-C. L{\'e}vesque, C.~Gagn{\'e}, and R.~Sabourin.
\newblock Bayesian hyperparameter optimization for ensemble learning.
\newblock In {\em Proceedings of the Thirty-Second Conference on Uncertainty in
  Artificial Intelligence}, pages 437--446, 2016.

\bibitem{li2020efficient}
Y.~Li, J.~Jiang, J.~Gao, Y.~Shao, C.~Zhang, and B.~Cui.
\newblock Efficient automatic cash via rising bandits.
\newblock In {\em Proceedings of the AAAI Conference on Artificial
  Intelligence}, volume~34, pages 4763--4771, 2020.

\bibitem{li2021mfes}
Y.~Li, Y.~Shen, J.~Jiang, J.~Gao, C.~Zhang, and B.~Cui.
\newblock Mfes-hb: Efficient hyperband with multi-fidelity quality
  measurements.
\newblock In {\em Proceedings of the AAAI Conference on Artificial
  Intelligence}, volume~35, pages 8491--8500, 2021.

\bibitem{li2021openbox}
Y.~Li, Y.~Shen, W.~Zhang, Y.~Chen, H.~Jiang, M.~Liu, J.~Jiang, J.~Gao, W.~Wu,
  Z.~Yang, C.~Zhang, and B.~Cui.
\newblock Openbox: A generalized black-box optimization service.
\newblock {\em Proceedings of the 27th ACM SIGKDD Conference on Knowledge
  Discovery \& Data Mining}, 2021.

\bibitem{li2021volcanoml}
Y.~Li, Y.~Shen, W.~Zhang, J.~Jiang, B.~Ding, Y.~Li, J.~Zhou, Z.~Yang, W.~Wu,
  C.~Zhang, et~al.
\newblock Volcanoml: speeding up end-to-end automl via scalable search space
  decomposition.
\newblock {\em Proceedings of the VLDB Endowment}, 14(11):2167--2176, 2021.

\bibitem{li2022volcanoml}
Y.~Li, Y.~Shen, W.~Zhang, C.~Zhang, and B.~Cui.
\newblock Efficient end-to-end automl via scalable search space decomposition.
\newblock {\em The VLDB Journal}, 2022.

\bibitem{limin2021balanced}
S.~Limin, Z.~Qiang, L.~Shuang, and L.~C. Harold.
\newblock Balanced discriminative transfer feature learning for visual domain
  adaptation.
\newblock {\em ZTE Communications}, 18(4):78--83, 2021.

\bibitem{liu2020value}
C.~Liu, B.~P. Chamberlain, and E.~J. McCoy.
\newblock What is the value of experimentation and measurement?
\newblock {\em Data Science and Engineering}, 5(2):152--167, 2020.

\bibitem{liu2020admm}
S.~Liu, P.~Ram, D.~Vijaykeerthy, D.~Bouneffouf, G.~Bramble, H.~Samulowitz,
  D.~Wang, A.~Conn, and A.~Gray.
\newblock An admm based framework for automl pipeline configuration.
\newblock In {\em Proceedings of the AAAI Conference on Artificial
  Intelligence}, volume~34, pages 4892--4899, 2020.

\bibitem{moghimi2016boosted}
M.~Moghimi, S.~J. Belongie, M.~J. Saberian, J.~Yang, N.~Vasconcelos, and L.-J.
  Li.
\newblock Boosted convolutional neural networks.
\newblock In {\em BMVC}, volume~5, page~6, 2016.

\bibitem{olson2016tpot}
R.~S. Olson and J.~H. Moore.
\newblock Tpot: A tree-based pipeline optimization tool for automating machine
  learning.
\newblock In {\em Workshop on automatic machine learning}, pages 66--74. PMLR,
  2016.

\bibitem{partalas2008focused}
I.~Partalas, G.~Tsoumakas, and I.~P. Vlahavas.
\newblock Focused ensemble selection: A diversity-based method for greedy
  ensemble selection.
\newblock In {\em ECAI}, pages 117--121, 2008.

\bibitem{quanming2018taking}
Y.~Quanming, W.~Mengshuo, J.~E. Hugo, G.~Isabelle, H.~Yi-Qi, L.~Yu-Feng,
  T.~Wei-Wei, Y.~Qiang, and Y.~Yang.
\newblock Taking human out of learning applications: A survey on automated
  machine learning.
\newblock {\em arXiv preprint arXiv:1810.13306}, 2018.

\bibitem{real2019regularized}
E.~Real, A.~Aggarwal, Y.~Huang, and Q.~V. Le.
\newblock Regularized evolution for image classifier architecture search.
\newblock In {\em Proceedings of the AAAI conference on artificial
  intelligence}, volume~33, pages 4780--4789, 2019.

\bibitem{snoek2012practical}
J.~Snoek, H.~Larochelle, and R.~P. Adams.
\newblock Practical bayesian optimization of machine learning algorithms.
\newblock In {\em Advances in neural information processing systems}, pages
  2951--2959, 2012.

\bibitem{su2009survey}
X.~Su and T.~M. Khoshgoftaar.
\newblock A survey of collaborative filtering techniques.
\newblock {\em Advances in artificial intelligence}, 2009, 2009.

\bibitem{sun2019bert4rec}
F.~Sun, J.~Liu, J.~Wu, C.~Pei, X.~Lin, W.~Ou, and P.~Jiang.
\newblock Bert4rec: Sequential recommendation with bidirectional encoder
  representations from transformer.
\newblock In {\em Proceedings of the 28th ACM international conference on
  information and knowledge management}, pages 1441--1450, 2019.

\bibitem{tang2006analysis}
E.~K. Tang, P.~N. Suganthan, and X.~Yao.
\newblock An analysis of diversity measures.
\newblock {\em Machine learning}, 65(1):247--271, 2006.

\bibitem{thornton2013auto}
C.~Thornton, F.~Hutter, H.~H. Hoos, and K.~Leyton-Brown.
\newblock Auto-weka: Combined selection and hyperparameter optimization of
  classification algorithms.
\newblock In {\em Proceedings of the 19th ACM SIGKDD international conference
  on Knowledge discovery and data mining}, pages 847--855, 2013.

\bibitem{vanschoren2014openml}
J.~Vanschoren, J.~N. Van~Rijn, B.~Bischl, and L.~Torgo.
\newblock Openml: networked science in machine learning.
\newblock {\em ACM SIGKDD Explorations Newsletter}, 15(2):49--60, 2014.

\bibitem{wang2021flaml}
C.~Wang, Q.~Wu, M.~Weimer, and E.~Zhu.
\newblock Flaml: a fast and lightweight automl library.
\newblock {\em Proceedings of Machine Learning and Systems}, 3:434--447, 2021.

\bibitem{yang2022diffusion}
L.~Yang, Z.~Zhang, and S.~Hong.
\newblock Diffusion models: A comprehensive survey of methods and applications.
\newblock {\em arXiv preprint arXiv:2209.00796}, 2022.

\bibitem{zaidi2021neural}
S.~Zaidi, A.~Zela, T.~Elsken, C.~C. Holmes, F.~Hutter, and Y.~Teh.
\newblock Neural ensemble search for uncertainty estimation and dataset shift.
\newblock {\em Advances in Neural Information Processing Systems}, 34, 2021.

\bibitem{zhang2020efficient}
W.~Zhang, J.~Jiang, Y.~Shao, and B.~Cui.
\newblock Efficient diversity-driven ensemble for deep neural networks.
\newblock In {\em 2020 IEEE 36th International Conference on Data Engineering
  (ICDE)}, pages 73--84. IEEE, 2020.

\bibitem{zhang2020snapshot}
W.~Zhang, J.~Jiang, Y.~Shao, and B.~Cui.
\newblock Snapshot boosting: a fast ensemble framework for deep neural
  networks.
\newblock {\em Science China Information Sciences}, 63(1):1--12, 2020.

\bibitem{zhang2022pasca}
W.~Zhang, Y.~Shen, Z.~Lin, Y.~Li, X.~Li, W.~Ouyang, Y.~Tao, Z.~Yang, and
  B.~Cui.
\newblock Pasca: A graph neural architecture search system under the scalable
  paradigm.
\newblock In {\em Proceedings of the ACM Web Conference 2022}, pages
  1817--1828, 2022.

\bibitem{zhou2012ensemble}
Z.-H. Zhou.
\newblock {\em Ensemble methods: foundations and algorithms}.
\newblock CRC press, 2012.

\bibitem{zhou2002ensembling}
Z.-H. Zhou, J.~Wu, and W.~Tang.
\newblock Ensembling neural networks: many could be better than all.
\newblock {\em Artificial intelligence}, 137(1-2):239--263, 2002.

\bibitem{zhu2021pre}
D.-H. Zhu, X.-Y. Dai, and J.-J. Chen.
\newblock Pre-train and learn: Preserving global information for graph neural
  networks.
\newblock {\em Journal of Computer Science and Technology}, 36(6):1420--1430,
  2021.

\bibitem{zimmer2021auto}
L.~Zimmer, M.~Lindauer, and F.~Hutter.
\newblock Auto-pytorch: multi-fidelity metalearning for efficient and robust
  autodl.
\newblock {\em IEEE Transactions on Pattern Analysis and Machine Intelligence},
  43(9):3079--3090, 2021.

\end{thebibliography}

\section*{Checklist}
\begin{enumerate}

\item For all authors...
\begin{enumerate}
  \item Do the main claims made in the abstract and introduction accurately reflect the paper's contributions and scope?
    \answerYes{}
  \item Did you describe the limitations of your work?
    \answerYes{See Section~\ref{sec:discussion}.}
  \item Did you discuss any potential negative societal impacts of your work?
    \answerNo
  \item Have you read the ethics review guidelines and ensured that your paper conforms to them?
    \answerYes{We have carefully read the guidelines.}
\end{enumerate}

\item If you are including theoretical results...
\begin{enumerate}
  \item Did you state the full set of assumptions of all theoretical results?
    \answerNA{}
	\item Did you include complete proofs of all theoretical results?
    \answerNA{}
\end{enumerate}

\item If you ran experiments...
\begin{enumerate}
  \item Did you include the code, data, and instructions needed to reproduce the main experimental results (either in the supplemental material or as a URL)?
    \answerYes{}
  \item Did you specify all the training details (e.g., data splits, hyperparameters, how they were chosen)?
    \answerYes{We have a detailed implementation instruction in Appendix A.4.}
	\item Did you report error bars (e.g., with respect to the random seed after running experiments multiple times)?
    \answerYes{See Table~\ref{tab:test} and Figure~\ref{fig:validation}.}
	\item Did you include the total amount of compute and the type of resources used (e.g., type of GPUs, internal cluster, or cloud provider)?
    \answerYes{See Appendix A.4.}
\end{enumerate}

\item If you are using existing assets (e.g., code, data, models) or curating/releasing new assets...
\begin{enumerate}
  \item If your work uses existing assets, did you cite the creators?
    \answerYes{}
  \item Did you mention the license of the assets?
    \answerNA{}
  \item Did you include any new assets either in the supplemental material or as a URL?
    \answerNA{}
  \item Did you discuss whether and how consent was obtained from people whose data you're using/curating?
    \answerYes{We use the widely used and public data on OpenML. The datasets are used for evaluation in previous work.}
  \item Did you discuss whether the data you are using/curating contains personally identifiable information or offensive content?
    \answerNo{}
\end{enumerate}

\item If you used crowdsourcing or conducted research with human subjects...
\begin{enumerate}
  \item Did you include the full text of instructions given to participants and screenshots, if applicable?
    \answerNA{}
  \item Did you describe any potential participant risks, with links to Institutional Review Board (IRB) approvals, if applicable?
    \answerNA{}
  \item Did you include the estimated hourly wage paid to participants and the total amount spent on participant compensation?
    \answerNA{}
\end{enumerate}

\end{enumerate}

\newpage
\appendix

\section{Appendix}
\subsection{Ensemble Selection}
\label{appendix:es}
We provide the pseudo-code in Algorithm~\ref{alg:framework}.
In our paper, the $Perf$ metric is the classification error based on the mean predictions of learners in the ensemble.

\begin{algorithm}[pbh]

  \caption{Procedure of ensemble selection.}
  \label{alg:framework}
    \KwIn{The ensemble size $E$, the configuration observations $D$, and the validation set $\mathcal{D}_{val}$.}
    Initialize the ensemble as $B=\varnothing$\;
    \For{$i=1,...,E$}{
     $a\leftarrow \mathop{\arg\min}_{a\in D}Perf(B\cup \{a\}, \mathcal{D}_{val})$\;
     $B \leftarrow B\cup \{a\}$\;
     }
    \textbf{return} the final ensemble $B$.
\end{algorithm}

\subsection{Dataset Details}
\label{appendix:dataset}
In Table~\ref{tab:dataset}, we provide the details of the datasets used in our experiment.
While all the datasets are collected from OpenML~\cite{vanschoren2014openml}, we provide the OpenML ID as the identification of the dataset.

\begin{table}[thb]
\centering
\resizebox{0.7\linewidth}{!}{
\begin{tabular}{lcccc}
    \toprule
    Datasets & OpenML ID & Classes & Samples & Features \\ 
    \midrule
    amazone\_employee & 43900 & 2 & 32769 & 9 \\
    bank32nh & 833 & 2 & 8192 & 32 \\
    cpu\_act & 761 & 2 & 8192 & 21  \\
    cpu\_small & 735 & 2 & 8192 & 12 \\
    eeg & 1471 & 2 & 14980 & 14 \\
    elevators & 846 & 2 & 16599 & 18 \\
    house\_8L & 843 & 2 & 22784 & 8 \\
    pol & 722 & 2 & 15000 & 48 \\
    pollen & 871 & 2 & 3848 & 5 \\
    puma32H & 752 & 2 & 8192 & 32 \\
    quake & 772 & 2 & 2178 & 3 \\
    satimage & 182 & 6 & 6430 & 36 \\
    spambase & 44 & 2 & 4600 & 57 \\
    wind & 847 & 2 & 6574 & 14 \\
    2dplanes & 727 & 2 & 40768 & 10\\
    \bottomrule
\end{tabular}
}
\caption{Basic dataset information.}
\label{tab:dataset}
\end{table}

\subsection{Search Space}
\label{appendix:search_space}
The search space for algorithms and feature engineering operators are presented in Tables~\ref{tab:space_algo} and ~\ref{tab:space_fe}, respectively.

\begin{minipage}[thb]{.5\textwidth}
  \centering
  \captionof{table}{Search space for algorithms. We distinguish categorical (cat) hyperparameters from numerical (cont) ones. The numbers in the brackets are conditional hyperparameters.}
  \resizebox{0.85\linewidth}{!}{
  \begin{tabular}{lccc}
    \toprule
    Type of Classifier & \#$\lambda$ & cat (cond) & cont (cond) \\ 
    \midrule
    AdaBoost & 4 & 1 (-) & 3 (-) \\
    Random forest & 5 & 2 (-) & 3 (-) \\
    Extra trees & 5 & 2 (-) & 3 (-) \\
    Gradient boosting & 7 & 1 (-) & 6 (-) \\
    KNN & 2 & 1 (-) & 1 (-) \\
    LDA & 4 & 1 (-) & 3 (1) \\
    QDA & 1 & - & 1 (-) \\
    Logistic regression & 4 & 2 (-) & 2 (-) \\
    Liblinear SVC & 5 & 2 (2) & 3 (-) \\
    LibSVM SVC & 7 & 2 (2) & 5 (-) \\
    LightGBM & 6 & - & 6 (-) \\
    \bottomrule
\end{tabular}}
  \label{tab:space_algo}
\end{minipage}\qquad
\begin{minipage}[tpb!]{.45\textwidth}
  \centering
  \captionof{table}{Search space of feature engineering operators.}
  \resizebox{0.88\linewidth}{!}{
  \begin{tabular}{p{120pt}ccc}
    \toprule
    Type of Operator & \#$\lambda$ & cat (cond) & cont (cond) \\ 
    \midrule
    Minmax & 0 & - & - \\
    Normalizer & 0 & - & - \\
    Quantile & 2 & 1 (-) & 1 (-) \\
    Robust & 2 & - & 2 (-) \\
    Standard & 0 & - & - \\
    \midrule
    Cross features & 1 & - & 1 (-) \\
    Fast ICA & 4 & 3 (1) & 1 (1) \\
    Feature agglomeration & 4 & 3 (2) & 1 (-) \\
    Kernel PCA & 5 & 1 (1) & 4 (3) \\
    Rand. kitchen sinks & 2 & - & 2 (-) \\
    LDA decomposer & 1 & 1 (-) & - \\
    Nystroem sampler & 5 & 1 (1) & 4 (3) \\
    PCA & 2 & 1 (-) & 1 (-) \\
    Polynomial & 2 & 1 (-) & 1 (-) \\
    Random trees embed. & 5 & 1 (-) & 4 (-) \\
    SVD & 1 & - & 1 (-) \\
    Select percentile & 2 & 1 (-) & 1 (-) \\
    Select generic univariate & 3 & 2 (-) & 1 (-) \\
    Extra trees preprocessing & 5 & 2 (-) & 3 (-) \\
    Linear SVM preprocessing & 5 & 3 (3) & 2 (-) \\
    \bottomrule
\end{tabular}}
  \label{tab:space_fe}

\end{minipage}

\subsection{Implementation Details}
\label{appendix:implementation}
We implement the performance surrogate of Bayesian optimization based on OpenBox~\cite{li2021openbox}, a toolkit for black-box optimization. 
The other baselines are implemented following the open-source version or original papers.
For NES, the population is set to 30; 
for EO, the ensemble size is set to 12;
for RB, $\alpha$ and trial per action are set to 3 and 5;
for BO and \sys, we sample 4950 and 50 candidates from global and local sampling, respectively;
for all post-hoc designs, we set the ensemble size of ensemble selection to 25;
for \sys, we set $\beta$ and $\tau$ to 0.05 and 0.2.
All the experiments are conducted on a machine with 64 `AMD EPYC 7702P' CPU cores.



\begin{figure*}[tp!]
\centering  
\subfigure[spambase]{
\includegraphics[width=0.45\textwidth]{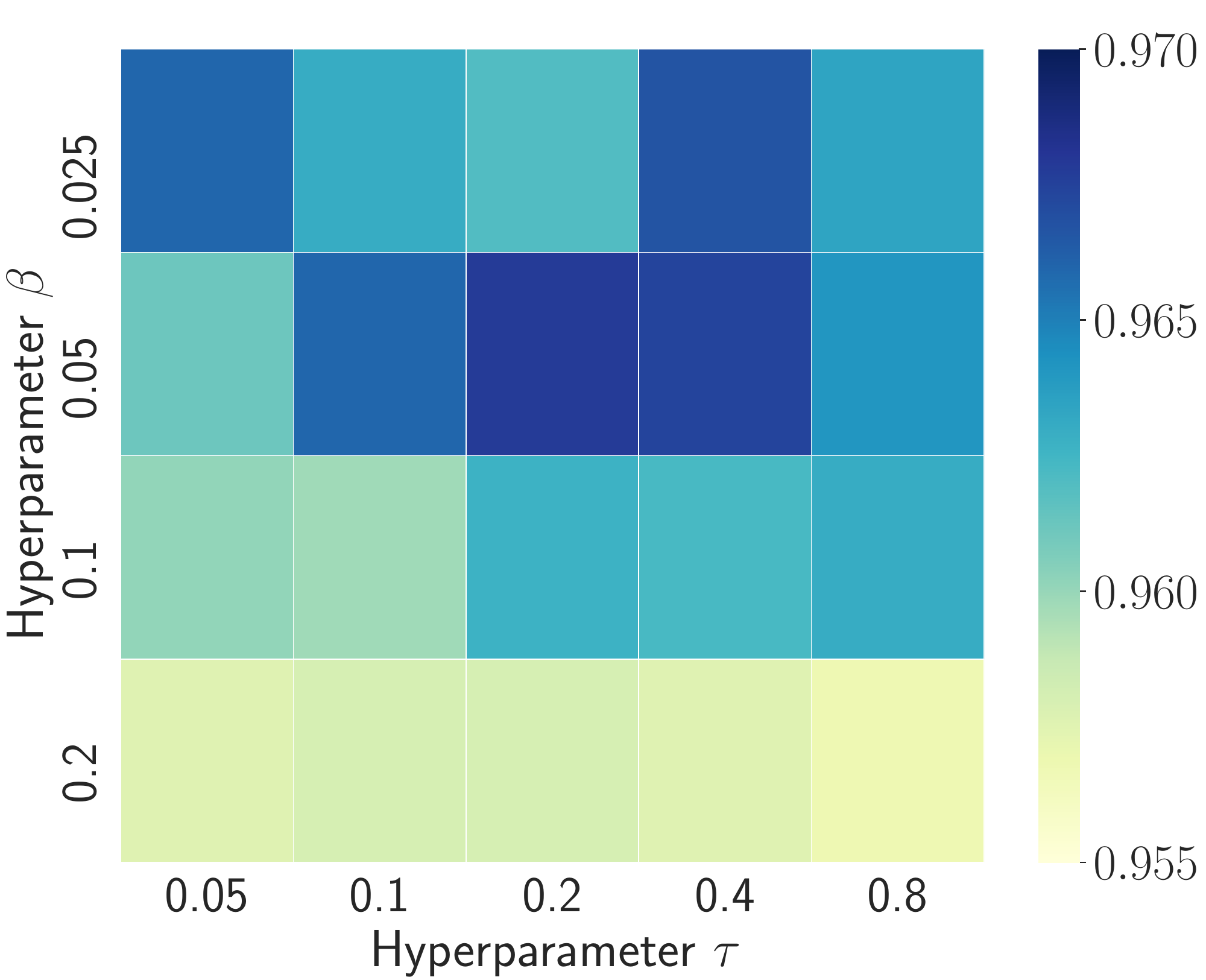}}\hspace{3mm}
\subfigure[house\_8L]{
\includegraphics[width=0.45\textwidth]{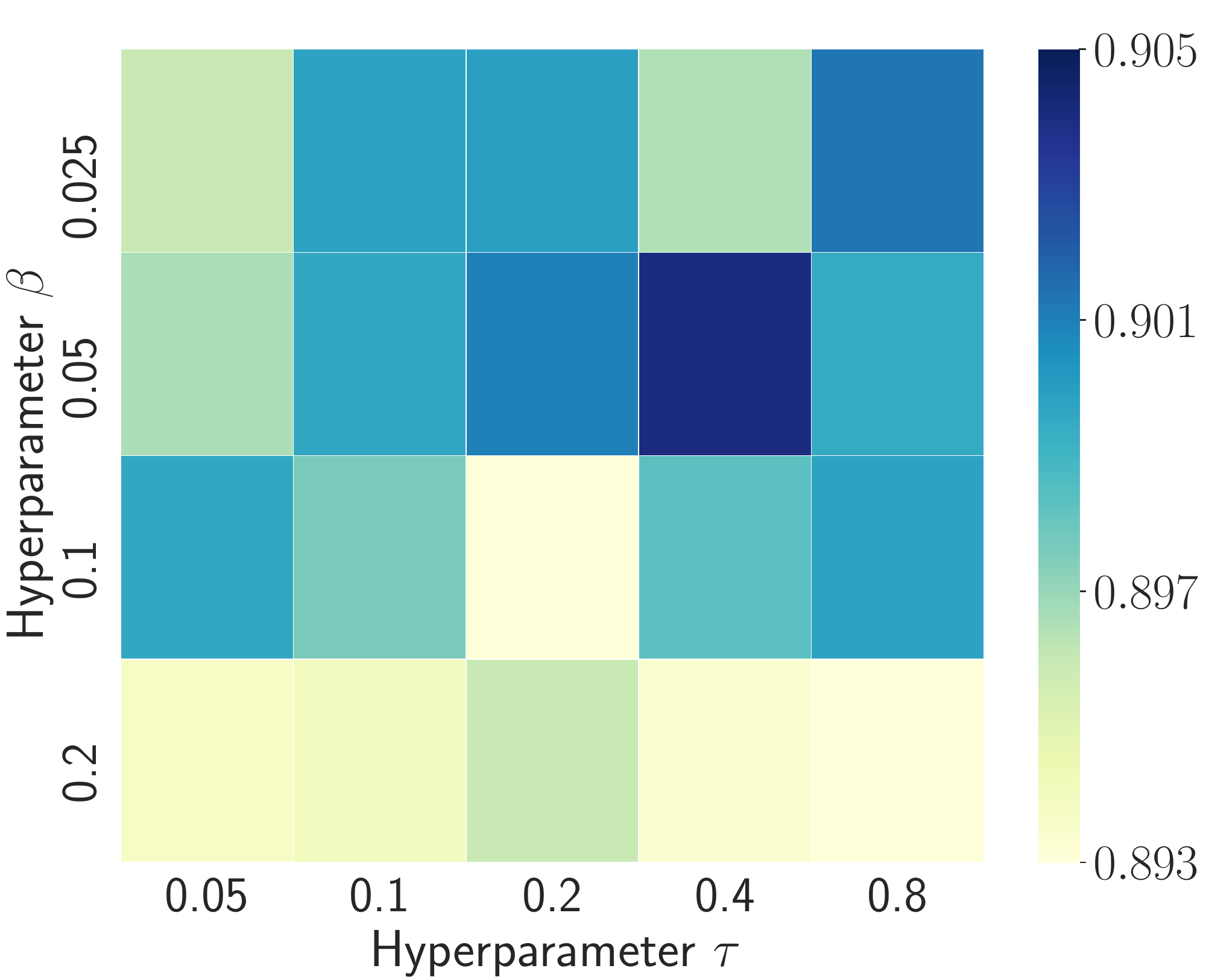}}
\caption{Sensitivity analysis on spambase and house\_8L.}
\label{fig:sensitivity}
\end{figure*}

\subsection{Additional Results}
\label{appendix:additional}
\textbf{Sensitivity Analysis.}
We first provide the hyperparameter sensitivity analysis on the dataset spambase.
The choices for $\beta$ are $\{0.025, 0.05, 0.1, 0.2\}$, and the choices for $\tau$ are
$\{0.05, 0.1, 0.2, 0.4\}$.
The validation accuracy of 20 combinations of hyperparameters are shown in Figure~\ref{fig:sensitivity}.
Remind that $\beta$ is the maximum of diversity importance and $\tau$ controls the speed of approaching saturation. We observe that a large $\beta$ (0.2) leads to a clear accuracy drop, and we suggest using a $\beta=0.05$. However, we need to tune $\tau$ to achieve the best results on different datasets. 
The reason may be that the difficulty for different datasets to find good configurations are different. 
As \sys builds on the intuition that we need to focus on accuracy rather than diversity in early iterations, a smaller $\tau$ is required if it's difficult to find accurate learners in early iterations. 
The suggested region for tuning $\tau$ is [0.1,0.8]. 
In our paper, we use 0.2 by default, but a tuned $\tau$ may achieve better results. 

\textbf{Ablation Study.}
In this part, we compare \sys without weight scheduling by setting $w=0.1$. The results on five datasets are shown in Table~\ref{tab:ablation}.
The results show that \sys with weight schedule (Equation 5) performs much better than fixing the weight for diversity (not significant on quake, but significant on other 4 datasets). 
It fits the intuition that motivates the weight schedule design in Section~\ref{sec:framework}.

\begin{table*}[thb]
\caption{Ablation study on weight scheduling.}
\vspace{-4mm}
\centering
\noindent
\resizebox{1.0\linewidth}{!}{
\subtable{
\begin{tabular}{cccccc}
\toprule
Method & elevators & house\_8L & pol & quake & wind \\
\midrule
\sys (fixed) & $9.59\pm0.30$ & $11.51\pm0.28$ & $1.66\pm0.19$ & $45.63\pm1.45$ & $14.27\pm0.36$\\
\sys & $\bm{9.40\pm0.28}$ & $\bm{10.80\pm0.22}$ & $\bm{1.34\pm0.17}$ & $\bm{45.55\pm1.37}$ & $\bm{13.93\pm0.42}$ \\
\bottomrule
\end{tabular}
}}
\label{tab:ablation}
\end{table*}

\textbf{Analysis on Ensemble Strategy.}
While previous study~\cite{feurer2015efficient} claims that ensemble selection performs well with CASH using Bayesian optimization, we also evaluate the influence of different strategies on \sys. 
We compare ensemble selection with weighted bagging and stacking.
We pick 5 learners with the best validation errors from observations to build ensemble for bagging and stacking following EO~\cite{levesque2016bayesian}.
The test results on 5 datasets are shown in Table~\ref{tab:ens_test}.
Among three compared strategies, ensemble selection performs the best.
In addition, the cost of bagging and ensemble selection can almost be ignored, since the intermediate predictions are stored during optimization. However, stacking is a time-consuming strategy, as the learners are re-trained multiple times after optimization. 
For example, if we pick 5 base learners for stacking and 5-fold cross validation is used, the cost of stacking equals to that of 25 configuration evaluations, while the budget for the entire CASH process is only 250 evaluations.
Therefore, we apply ensemble selection as the default post-hoc ensemble strategy in our paper.

\begin{table*}[thb]
\caption{Test error (\%) of different ensemble strategies on different datasets.}
\vspace{-4mm}
\centering
\noindent
\resizebox{1.0\linewidth}{!}{
\subtable{
\begin{tabular}{cccccc}
\toprule
Method & elevators & house\_8L & pol & quake & wind \\
\midrule
Bagging & $10.98\pm1.70$ & $11.12\pm0.33$ & $1.85\pm0.31$ & $47.22\pm1.85$ & $14.21\pm2.82$\\
Stacking & $9.53\pm1.13$ & $11.04\pm0.23$ & $1.54\pm0.28$ & $46.58\pm1.13$ & $14.02\pm1.43$\\
Ensemble selection & $\bm{9.40\pm0.28}$ & $\bm{10.80\pm0.22}$ & $\bm{1.34\pm0.17}$ & $\bm{45.55\pm1.37}$ & $\bm{13.93\pm0.42}$ \\
\bottomrule
\end{tabular}
}}
\label{tab:ens_test}
\end{table*}

\textbf{Analysis on Ensemble Size.}
We also provide the results if we set a larger ensemble size. 
As ensemble selection directly optimizes the performance on the validation set, the validation performance is definitely no worse than using a smaller ensemble size due to the greedy mechanism. 
However, as pointed out by~\cite{hutter2019automated}, if we optimize the validation set too much (i.e., setting a too large ensemble size for ensemble selection), the test results may deteriorate, which is referred to as the overfitting issue in AutoML. 
The results when setting the ensemble size to 100 for BO-ES are presented in Table~\ref{tab:ens_size}.

\begin{table*}[thb]
\caption{Test error (\%) of different ensemble sizes on different datasets.}
\vspace{-4mm}
\centering
\noindent
\resizebox{1.0\linewidth}{!}{
\subtable{
\begin{tabular}{cccccc}
\toprule
Method & elevators & house\_8L & pol & quake & wind \\
\midrule
BO-ES (ens\_size=100) & $9.98\pm0.30$ & $11.52\pm0.26$ & $1.45\pm0.33$ & $47.43\pm1.62$ & $14.04\pm0.47$\\
BO-ES (ens\_size=25) & $\bm{9.61\pm0.36}$ & $\bm{11.06\pm0.33}$ & $\bm{1.35\pm0.18}$ & $\bm{46.10\pm2.52}$ & $14.04\pm0.53$\\
\bottomrule
\end{tabular}
}}
\label{tab:ens_size}
\end{table*}

In main paper, the ensemble size is set to 25 following VolcanoML~\cite{li2021volcanoml}, which shows good empirical results across different datasets. 
We observe that when we set the ensemble size to 100 for BO-ES, the test results are generally worse than setting the ensemble size to 25 due to the overfitting issue (not significant on wind but significant on the other four). 
We have also mentioned this risk of overfitting in the limitation.

\textbf{Comparison with Intuitive Designs.}
In this part, we evaluate another intuitive CASH design, which tunes each algorithm for the same budget and then builds a post-hoc ensemble. 
In fact, it is a simplified version of the baseline RB-ES, in which RB-ES eliminates some of the algorithms after several iterations. 
We name it kBO-ES and present the results in Table~\ref{tab:kbo_es}.
We observe that the results of kBO-ES are quite similar to RS-ES (Random search with ensemble selection). 
The reason is that the search space contains a lot of algorithms while the budget is quite limited (250 iterations). Each algorithm can only be tuned about 22 times. 
For each algorithm, we also need to tune the feature engineering operators (>50 HPs in auto-sklearn search space), and thus the BO surrogate for each algorithm is under-fitted. 
Therefore, Bayesian optimization for each algorithm performs like random search. 
kBO-ES is an intuitive method but seems to perform not competitively when the search space is very large.

\begin{table*}[thb]
\caption{Test error (\%) compared with another intuitive design.}
\vspace{-4mm}
\centering
\noindent
\resizebox{1.0\linewidth}{!}{
\subtable{
\begin{tabular}{cccccc}
\toprule
Method & elevators & house\_8L & pol & quake & wind \\
\midrule
RS-ES & $9.51\pm0.28$ & $11.21\pm0.38$ & $1.39\pm0.15$ & $46.79\pm1.57$ & $14.34\pm0.47$\\
kBO-ES& $9.55\pm0.32$ & $11.18\pm0.34$ & $1.39\pm0.16$ & $46.81\pm1.48$ & $14.29\pm0.45$\\
\sys & $\bm{9.40\pm0.28}$ & $\bm{10.80\pm0.22}$ & $\bm{1.34\pm0.17}$ & $\bm{45.55\pm1.37}$ & $\bm{13.93\pm0.42}$ \\
\bottomrule
\end{tabular}
}}
\label{tab:kbo_es}
\end{table*}

\textbf{Comparison with AutoGluon.}
The search space plays a significant role in CASH optimization. 
As \sys is an algorithm framework rather than a system, we compare it with other baselines by using the same search space (i.e., auto-sklearn space). 
However, AutoGluon applies a more compact space than auto-sklearn, and it's not fair to directly compare DivBO on auto-sklearn search space with AutoGluon. 
To make a relatively fair comparison, we reproduce a similar search space of AutoGluon except for the specified neural networks due to implementation difficulty. The results on five datasets are in Table~\ref{tab:autogluon}.

\begin{table*}[thb]
\caption{Test error (\%) compared with AutoGluon.}
\vspace{-4mm}
\centering
\noindent
\resizebox{1.0\linewidth}{!}{
\subtable{
\begin{tabular}{cccccc}
\toprule
Method & elevators & house\_8L & pol & quake & wind \\
\midrule
AutoGluon Tabular & $9.10\pm0.00$ & $\bm{9.98\pm0.00}$ & $1.23\pm0.00$ & $44.72\pm0.00$ & $14.37\pm0.00$\\
\sys (AutoGluon) & $\bm{9.01\pm0.11}$ & $10.06\pm0.17$ & $\bm{1.18\pm0.07}$ & $44.75\pm0.60$ & \bm{$14.24\pm0.18$}\\
\bottomrule
\end{tabular}
}}
\label{tab:autogluon}
\end{table*}

Note that, the search space affects the results a lot. For example, AutoGluon's results on wind are worse than RS-ES using the auto-sklearn space. 
However, AutoGluon's results on the other four datasets are better than most of the results using the auto-sklearn space, which is consistent with the observation that AutoGluon often outperforms auto-sklearn. 
The reason may be that AutoGluon is equipped with a well-designed search space, which kicks out less reliable algorithms on modern datasets (e.g., Naive Bayes) and adds strong ones (e.g., Catboost). 
The variance of AutoGluon's results is zero because it fixes the random seed in its inner design. 
In addition, we observe an error decrease when using DivBO in this search space. Concretely, the improvement is statistically significant on three datasets, not significant on one (quake), and slightly worse on the other one (house\_8L).



\end{document}